\documentclass[twoside,11pt]{article}

\usepackage{blindtext}

\usepackage{jmlr2e}
\usepackage{wrapfig}
\usepackage{lscape}

\usepackage{algorithm} 
\usepackage{algorithmic}
\usepackage{colortbl, fontawesome, soul}
\usepackage{graphicx}
\usepackage{subcaption}
\usepackage{array}
\usepackage[export]{adjustbox}

\usepackage[utf8]{inputenc} 
\usepackage[T1]{fontenc}    
\usepackage{hyperref}       
\usepackage{url}            
\usepackage{booktabs}       
\usepackage{amsfonts}       
\usepackage{nicefrac}       
\usepackage{microtype}      
\usepackage{xcolor}         
\usepackage{amsmath, amssymb}
\newcolumntype{H}{>{\setbox0=\hbox\bgroup}c<{\egroup}@{}}

\usepackage{bm}
\usepackage{graphicx,wrapfig}
\usepackage{subcaption}
\usepackage{tabularray}
\usepackage{listings}
\usepackage{hhline}
\usepackage{listings}
\usepackage{multirow}
\usepackage{float}
\usepackage{algorithm} 
\usepackage{algorithmic}
\usepackage{cleveref}

\usepackage{enumitem}
\usepackage{colortbl, fontawesome, soul, mathtools,pifont}

\usepackage{lastpage}
\jmlrheading{23}{2022}{1-\pageref{LastPage}}{1/21; Revised 5/22}{9/22}{21-0000}{Author One and Author Two}

\ShortHeadings{On the Use of Anchoring for Training Vision Models}{Preprint}
\firstpageno{1}

\begin{document}

\title{On the Use of Anchoring for Training Vision Models} 

\author{\name Vivek Narayanaswamy \email narayanaswam1@llnl.gov \\
       \addr Lawrence Livermore National Laboratory 
       \AND
       \name Kowshik Thopalli \email thopalli1@llnl.gov \\
       \addr Lawrence Livermore National Laboratory
       \AND
       \name Rushil Anirudh \email rushil15anirudh@gmail.com \\
       \addr Amazon 
       \AND
       \name Yamen Mubarka \email mubarkba1@llnl.gov \\
       \addr Lawrence Livermore National Laboratory 
       \AND
       \name Wesam Sakla \email sakla1@llnll.gov \\
       \addr Lawrence Livermore National Laboratory
       \AND
       \name Jayaraman J. Thiagarajan \email jjayaram@llnl.gov \\
       \addr Lawrence Livermore National Laboratory 
       }

\editor{}

\maketitle

\begin{abstract}
Anchoring is a recent, architecture-agnostic principle for training deep neural networks that has been shown to significantly improve uncertainty estimation, calibration, and extrapolation capabilities. In this paper, we systematically explore anchoring as a general protocol for training vision models, providing fundamental insights into its training and inference processes and their implications for generalization and safety. Despite its promise, we identify a critical problem in anchored training that can lead to an increased risk of learning undesirable shortcuts, thereby limiting its generalization capabilities. To address this, we introduce a new anchored training protocol that employs a simple regularizer to mitigate this issue and significantly enhances generalization. We empirically evaluate our proposed approach across datasets and architectures of varying scales and complexities, demonstrating substantial performance gains in generalization and safety metrics compared to the standard training protocol.

\end{abstract}


\section{Introduction}
Anchoring~\citep{thiagarajan2022single} is a recent architecture-agnostic principle for training deep neural networks. It reparameterizes each input $\mathrm{x}$ into a tuple comprising a reference sample $\bar{\mathrm{r}}$ and the \textit{residual} $\mathrm{d} = \mathrm{x}-\bar{\mathrm{r}}$, i.e., $[\bar{\mathrm{r}}, \mathrm{d}]$, $\bar{\mathrm{r}} \sim P_{\mathrm{r}}$ and $\mathrm{d} \sim P_{\Delta}$. Here, $P_{\mathrm{r}}$ and $P_{\Delta}$ denote the distributions of references and residuals respectively. The resulting tuple is then fed as input to a deep network instead of the original input $\mathrm{x}$, by concatenating the tuple elements along the feature axis for vector-valued data or the channel axis for image data. Although the first layer of the network needs to be modified to accommodate twice the number of input dimensions (due to concatenation), the rest of the model architecture and optimization strategies remain the same as in standard training. This simple re-parameterization of the input forces the neural network to model the joint distribution $P_{(\mathrm{r}, \Delta)}$ for predicting the target label $\mathrm{y}$. Formally, the training objective can be written as:
\begin{equation}
    \theta^* = \arg \underset{\theta}{\min}~~~\frac{1}{|\mathcal{D}|} \sum_{(\mathrm{x}, \mathrm{y}) \in \mathcal{D}} \underset{\bar{\mathrm{r}}\sim P_{\mathrm{r}}}{\mathbb{E}} \mathcal{L}\bigg[\mathrm{y}, \mathcal{F}_\theta\bigg(\texttt{concat}([\bar{\mathrm{r}}, \mathrm{x}-\bar{\mathrm{r}}])\bigg)\bigg], 
\label{eqn:main}
\end{equation}
where $\mathcal{L}(.)$ is a loss function such as cross-entropy, $\mathcal{D}$ is the training dataset and $\mathcal{F}$ is the underlying network parameterized by $\theta$. In effect, for a given $\mathrm{x}$ and reference samples $\bar{\mathrm{r}}_1, \dots, \bar{\mathrm{r}}_k$, anchoring ensures that $\mathcal{F}_\theta([\bar{\mathrm{r}}_1, \mathrm{d}_1]) = \dots =\mathcal{F}_\theta([\bar{\mathrm{r}}_k,\mathrm{d}_k])$, where $\mathrm{d}_k = \mathrm{x}-\bar{\mathrm{r}}_k$. In other words, regardless of the choice of reference the model must arrive at the same prediction for an input. This principle has been shown to produce models with improved calibration and extrapolation properties~\citep{netanyahu2023learning, anirudh2023out}, and to facilitate accurate epistemic uncertainty estimation~\citep{thiagarajan2022single}. In this paper, we systematically explore the utility of anchoring as a generic protocol for building vision models and make a number of fundamental insights on its training and inferencing, applicability to different architecture families (conv-nets, transformers), and most importantly, the implications on model generalization and safety.

Our main contributions in this work can be summarized as follows:

\noindent \textbf{A closer look into anchored training and inferencing}: By studying the roles of reference set diversity and the inferencing protocol choice on the behavior of anchored models, we identify a critical limitation in current practice. More specifically, we find that conventional anchored training fails to effectively leverage the reference diversity, thus restricting its generalization capabilities, and that merely adopting sophisticated inference protocols~\cite{netanyahu2023learning} cannot circumvent this limitation.

\noindent \textbf{A new anchored training protocol}: We attribute the limited generalization power of anchored models to the increased risk of learning undesirable shortcuts, owing to insufficient sampling of $P_{(\mathrm{r}, \Delta)}$ during training, particularly in cases of high reference diversity. To address this, we introduce a new training protocol for anchoring that relies on a novel reference-masking regularizer.

\noindent \textbf{Benchmarking generalization and safety of anchored models}: Since anchoring is architecture-agnostic, we benchmark it using a variety of conv-net/transformer architectures on CIFAR-10, CIFAR-100 and Imagenet-1K datasets. We demonstrate significant improvements in OOD generalization, calibration and anomaly resilience over standard training. We also show that, without incurring any additional training or inference overheads, anchoring is synergistic to existing training strategies (e.g., data augmentations, optimizers, schedulers).

\section{A Closer Look into Anchored Training and Inference}
\subsection{What makes anchoring a promising training protocol?} 
Anchored training forces the network to learn a mapping between the joint space of (reference, residuals) and the targets, rather than the original input-target pairs. At first glance, anchoring may seem like a trivial reposing of standard training, but it is conceptually very different. Through this reparameterization, anchoring creates different relative representations for a sample with respect to references drawn from $P_\mathrm{r}$, and attempts to marginalize the effect of the reference when making a prediction for that sample. As demonstrated by~\citep{thiagarajan2022single}, this process exploits the lack of shift invariance in the neural tangent kernel induced by deep networks~\citep{jacot2018neural}, and implicitly explores a wider hypothesis class that is potentially more generalizable. Furthermore, anchored models have been found to extrapolate better to unseen data regimes through the use of transductive inferencing~\citep{netanyahu2023learning}, i.e., identifying an optimal reference for each sample, such that the resulting residual is likely to have been exposed to the model during training. While anchoring offers promise, its success hinges on effectively leveraging the diversity of the reference-residual pairs and stably converging for the same protocols from standard training (e.g., architectures, data augmentations, optimizers etc.).

\subsection{Does reference diversity play a key role in anchored training ?}
A unique property of anchoring is its ability to utilize relative representations w.r.t. a reference distribution $P_\mathrm{r}$ (realized using a reference set $\mathcal{R}$), effectively operating in the joint space $P_{(\mathrm{r}, \Delta)}$. During implementation, the reference set $\mathcal{R}$ is defined as a subset of the training data itself i.e., $\mathcal{R} \subseteq \mathcal{D}$~\citep{thiagarajan2022single}. Intuitively, by controlling the construction of $\mathcal{R}$, one can control the diversity of reference-residual combinations that anchored training is exposed to. We hope that with exposure to increasingly large and diverse reference sets, anchoring will explore a wide range of hypotheses, while also ensuring that the model can make consistent predictions for test samples using any randomly drawn reference $\bar{\mathrm{r}} \in \mathcal{R}$. However, when the anchored training does not effectively characterize the joint distribution $P_{(\mathrm{r}, \Delta)}$, the generalization can suffer, particularly when tested beyond the regimes of training data. To obtain a deeper understanding of anchored training, we conduct an empirical study on CIFAR10/100 datasets by varying the diversity of $\mathcal{R}$.

\noindent\underline{\textit{Setup}}. We first sub-sample $\mathcal{D}$ to construct reference sets of varying sizes ranging between 5 and 50000, where the latter corresponds to the entire training dataset. The construction is such that each set represents an increasing level of sample diversity (i.e., samples from multiple classes). This is followed by anchored training based on the different reference sets with ResNet18 models~\citep{he2016deep}. All other training specifics and hyper-parameters are fixed across the experiments. Post-training, we evaluate the model performance on the CIFAR10C/100C synthetic corruption benchmarks~\citep{hendrycks2018benchmarking} and report the average corruption accuracy across $5$ corruption severity levels.

\begin{figure}[t]
    \centering
    \begin{subfigure}{0.45\linewidth}
        \centering
        \includegraphics[width=\linewidth]{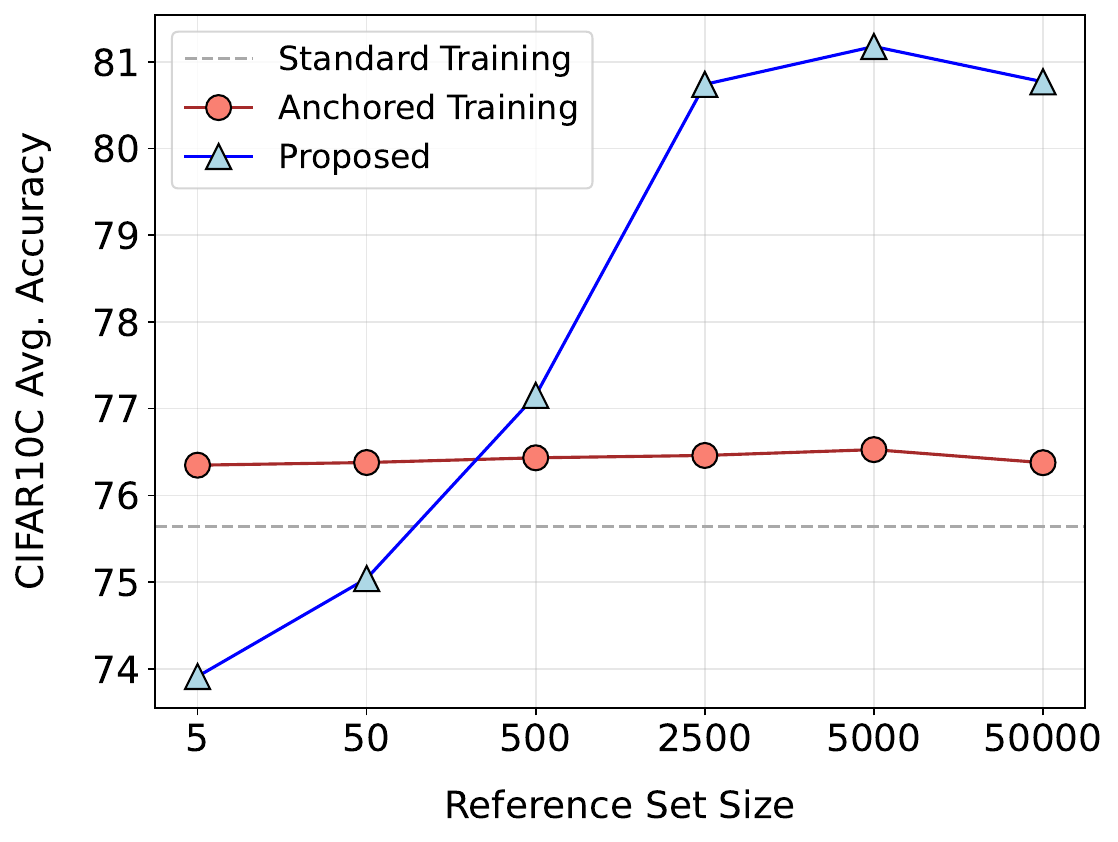}
        \caption{}
        \label{fig:anchorset_cifar10}
    \end{subfigure}
    \hfill
    \begin{subfigure}{0.45\linewidth}
        \centering
        \includegraphics[width=\linewidth]{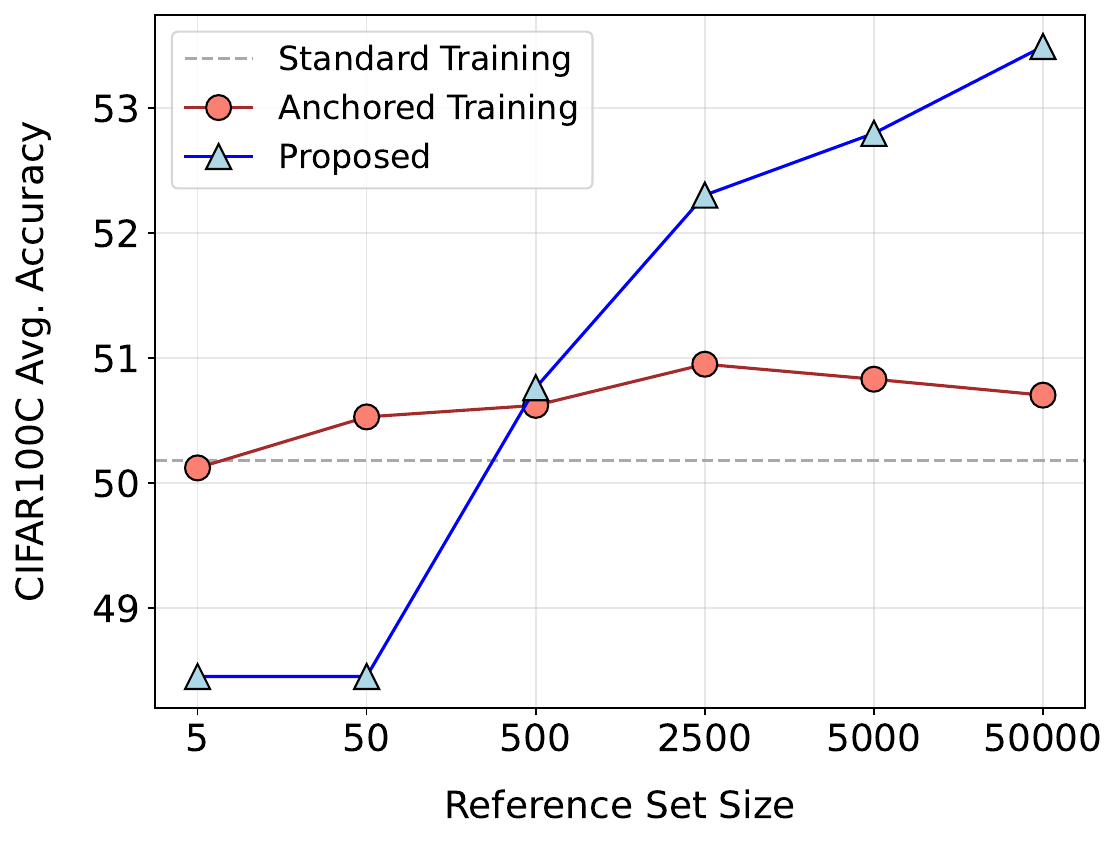}
        \caption{}
        \label{fig:anchorset_cifar100}
    \end{subfigure}
    \caption{\textbf{Impact of reference set size on anchored training performance}. With increase in reference set size, anchoring explores more diverse combinations of reference-residual pairs with the hope of demonstrating improved generalization performance. Surprisingly, the existing anchored training protocol does not effectively leverage this diversity even with increased reference set size albeit providing improvements in accuracy over standard training. We propose reference masking, a simple regularization strategy for training anchored models that recovers the lost performance.}
    \label{fig:anchorset}
\end{figure}

\noindent\underline{\textit{Observations}}. \Cref{fig:anchorset_cifar10} and \ref{fig:anchorset_cifar100} illustrates the performance of CIFAR10/100 anchored training on the respective evaluation benchmarks. Interestingly, we observe that the anchoring performance remains fairly similar (minor improvements in accuracy) even with orders of magnitude growth in the reference set size. While anchoring provides consistent benefits over standard training ($0.5\%-1\%$ on average), it is clear that the growing diversity of $P_{(\mathrm{r}, \Delta)}$ is not fully leveraged. \textit{This observation is in contrary to the insights from existing works, which recommend the use of the entire train data as the reference set for maximal benefits}. It is also worth noting that we utilize a single random reference (from the respective sets) to perform inference. This naturally raises the question if a more sophisticated inference protocol circumvent this limitation that we notice in anchored models.

\subsection{Can the choice of inference protocol improve the performance of anchored models?}
From existing works on anchoring, we find that different inference protocols can be used to elicit improvements in uncertainty quantification and model extrapolation. For instance, Thiagarajan \textit{et al.}~\citep{thiagarajan2022single} employed a reference marginalization strategy that samples $K$ random references from the reference set to obtain $K$ independent predictions for a given input (similar to MC-dropout or deep ensembles). This is followed by computing the prediction average along with its standard deviation, wherein the latter was interpreted as an estimate of epistemic uncertainty. The intuition is that different reference-residual combinations can lead to slightly different predictions for test sample that has not been observed during training, and marginalizing across references can offer robustness. On the other hand, Netanyahu \textit{et al.}~\citep{netanyahu2023learning} introduced the bilinear transduction (BLT) protocol for performing extrapolation from unseen data regimes in regression tasks. It was found that generalizing to an ``out of support'' (OOS) sample $\mathrm{x}_t$ (i.e., no evidence of observing such a sample in the training data) can be made more tractable by carefully choosing anchors $\Tilde{\mathrm{r}}\sim P_\mathrm{r}$ such that $\mathrm{x}_t-\tilde{\mathrm{r}} = \tilde{\mathrm{d}} \sim P_\Delta$. It was argued that, even if the specific combination of $[\Tilde{\mathrm{r}},\mathrm{x}_t-\tilde{\mathrm{r}}]$ may not be observed during training, the anchored model can produce better calibrated predictions when $\tilde{\mathrm{r}} \in P_{\mathrm{r}}$ and $\tilde{\mathrm{d}} \sim P_\Delta$. This is in contrast to~\cite{thiagarajan2022single}, which hypothesized that when the tuple $[\tilde{\mathrm{r}}, \mathrm{x}_t - \tilde{\mathrm{r}}] \notin P_{(\mathrm{r}, \Delta)}$, the inconsistency in the prediction will manifest as epistemic uncertainties. However, neither of these clearly answer the impact of inference protocol choice on generalization performance, particularly when the reference set diversity is high. To answer this, we conducted a systematic evaluation of these protocols with anchored models trained on CIFAR100 with the reference set $\mathcal{R} = \mathcal{D}$.

\noindent\underline{\textit{Setup}}. We consider three evaluation protocols to make predictions for the CIFAR100C benchmark (i) $1$ Random, that utilizes a single reference (e.g., average of samples in $\mathcal{R}$) to obtain predictions; (ii) $K$ Random that utilizes $K$ random references followed by reference marginalization ($K=10$ in our case); (iii) BLT that searches for the optimal reference in $\mathcal{R}$ for each test sample. Since conducting such an exhaustive search can be expensive for bigger datasets, we pick a subset (set to $50$ in our experiment).

\begin{figure}[t]
    \centering
    \includegraphics[width=0.99\linewidth]{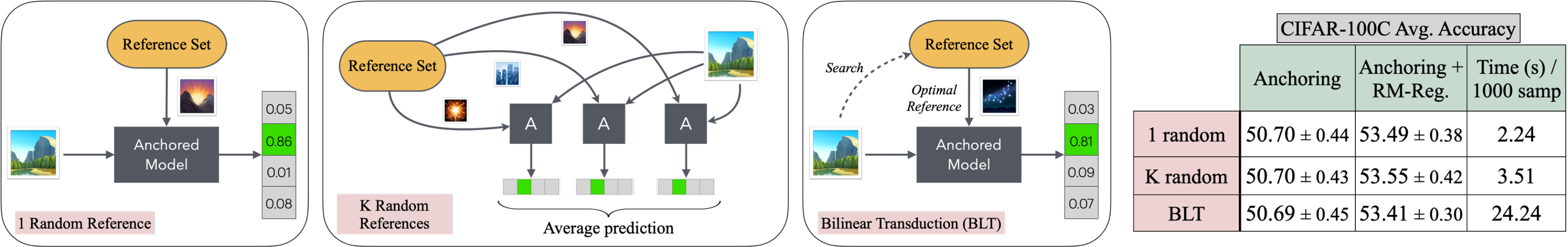}
    \caption{\textbf{Impact of the choice of inference protocol on the performance of anchored models~\citep{thiagarajan2022single, netanyahu2023learning}.} (Left) A single random reference is chosen for sample prediction; (Middle) Obtaining predictions using K random references followed by averaging; (Right) Bilinear Transduction that identifies the optimal reference for each sample. We find that, while these protocols have varying computational complexities (time (s)/1000 samples), there are no apparent gaps in the performance, indicating that the limitation of anchored training cannot be fixed through sophisticated inference protocols. }
    \label{fig:inference}
\end{figure}

\noindent\underline{\textit{Observations}}. The table in \Cref{fig:inference} provides the average accuracies obtained from these inference protocols. Interestingly, while these protocols incurs varying inference times (column 3) (BLT $>>$ $K$ random $>$ $1$ random), their accuracies are statistically similar to each other (averaged across multiple seeds). \textit{This observation implies that that the limitation of anchored training cannot be fixed through sophisticated inference protocols}. This motivates us to revisit anchoring and investigate if its behavior can be systematically improved during training itself.

\section{Improving Anchored Training via Reference Masking Regularization}
\begin{wrapfigure}{c}{0.4\textwidth}
    \centering
    \includegraphics[width=0.38\textwidth]{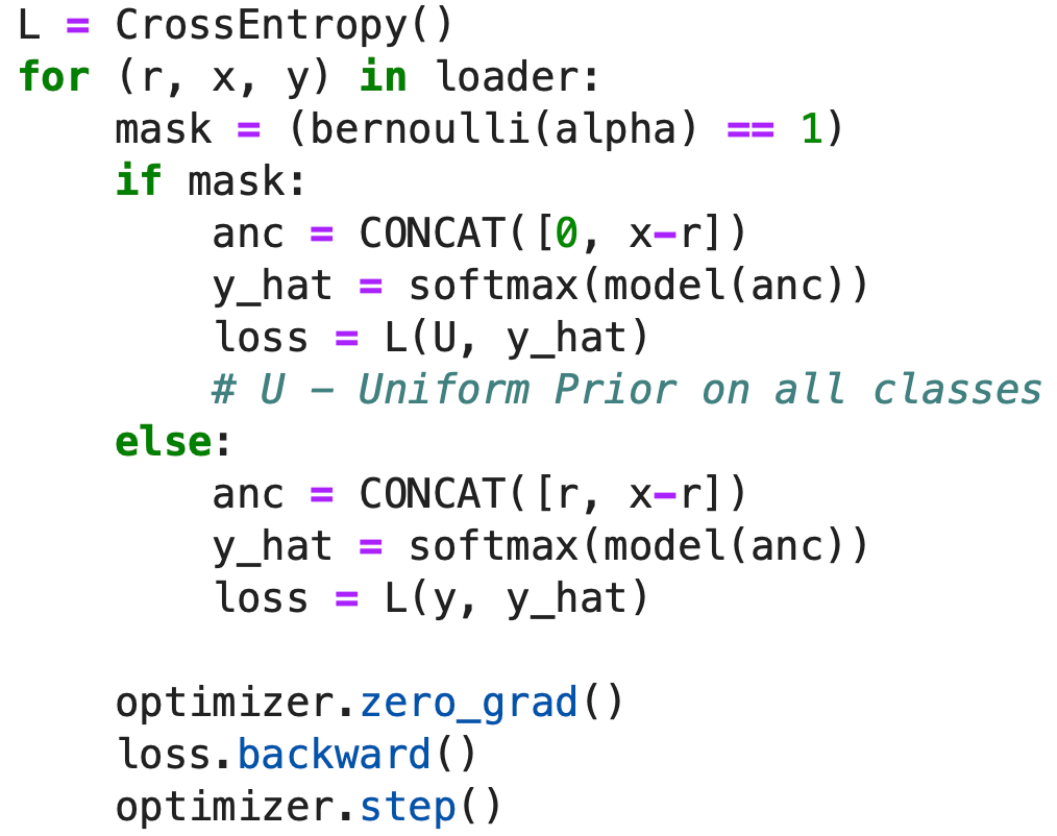} 
    \caption{{PyTorch style pseudo code for our proposed approach}.}
    \label{fig:algo}
\end{wrapfigure}

A close examination of anchored training reveals a critical limitation. As the size of the reference set increases, the number of reference-residual pairs grows combinatorially. For example, when $\mathcal{R} = \mathcal{D}$, there are ${|\mathcal{R}| \choose 2}$ possible pairs, making it impractical to explore all pairs within a fixed number of training iterations. This results in insufficient sampling of $P_{(\mathrm{r},\Delta)}$, increasing the risk that anchored training may overlook the reference and make predictions based solely on the residuals. Such non-generalizable shortcuts are problematic because a sample should not be identifiable without considering the reference. Therefore, it is crucial to enhance anchored training by more effectively utilizing the diversity present in large reference sets.

\subsection{Reference Masking Regularization}
We propose a novel, yet simple regularization strategy for improving anchored training. Formally, for a given tuple $[\bar{\mathrm{r}}, \mathrm{x} - \bar{\mathrm{r}}]$, and a user specified probability $\alpha$ that controls how often the training is regularized, reference masking zeroes out the reference and keeps the residual fixed to obtain $[\mathbf{0}, \mathrm{x}-\bar{\mathrm{r}}]$. For comparison, the tuple for the same sample $\mathrm{x}$ but with a ``zero'' reference (Note: zero vector/image can be a valid reference in our reference distribution) corresponds to $[\mathbf{0}, \mathrm{x} - \mathbf{0}]$. In order to preserve the integrity of the anchoring mechanism, we systematically discourage the model from making meaningful predictions when the reference is masked. This can be implemented by mapping randomly masked tuples to high-entropy predictions (i.e., uniform probabilities). We achieve this by minimizing the cross-entropy loss between the predictions from the masked tuple and the uniform prior $\mathcal{U}$ over $C$ classes (i.e, probability of any class $=1/C$). \Cref{fig:algo} provides the algorithm our proposed approach. 

Circling back to \Cref{fig:anchorset}, we observe that the proposed regularization significantly improves generalization accuracies compared to standard and original anchored training. This clearly demonstrates our strategy's effectiveness in leveraging the diversity in $P_{{\mathrm{r}}, \Delta}$. Following the insights from the previous section, we use the simple $1$ random inferencing protocol to obtain predictions for test samples. At low anchor set sizes ($|\mathcal{R}|\leq 50$), there is high likelihood of exposing the model to all possible combinations of samples and references, and hence the risk of learning such shortcuts is minimal. In such a scenario, overemphasizing the masking-based regularization (i.e., high $\alpha$) leads to underfitting, as illustrated in \Cref{fig:anchorset}. Unsurprisingly, reducing the masking probability can circumvent this underfitting behavior, as evidenced by the original anchored training, where $\alpha=0$. However, the benefits of our regularization become apparent at larger reference set sizes. Additionally, the table in \Cref{fig:inference} demonstrates that our approach performs similarly to the original anchored training, thereby implying no discernible impact on the inference efficiency.
\begin{figure}[t]
    \centering
    \includegraphics[width=\linewidth]{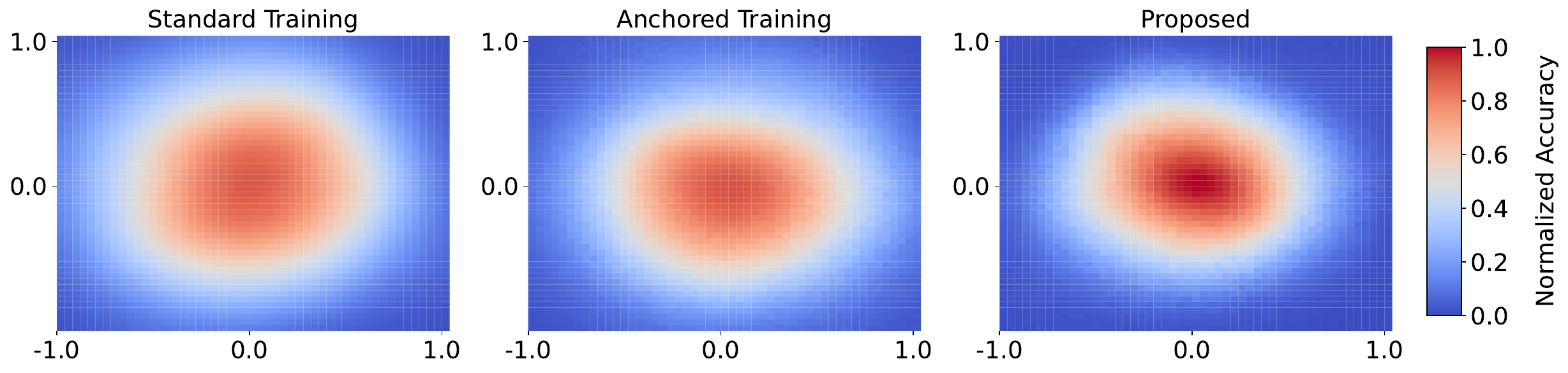}
    \caption{\textbf{Impact of the proposed regularizer on anchored training.} Using the CIFAR100C accuracy landscape, i.e., 2D heatmaps of the parameter space, we find that our approach identifies flatter and wider optima, thus leading to improved generalization~\cite{garipov2018loss}}
    \label{fig:landscape}
\end{figure}

\begin{figure}[ht]
    \centering
    \begin{subfigure}[c]{0.58\textwidth}
        \centering
        \renewcommand{\arraystretch}{1.8}
        \resizebox{\linewidth}{!}{
        \begin{tabular}{|c|c|c|c|}
        \hline
        \rowcolor[HTML]{EFEFEF} 
        \textbf{Augmentations} & \textbf{Standard Training} & \textbf{Anchored Training} & \textbf{Proposed} \\ \hline
        Geometric   & 33.98 & 34.43 &  \textbf{38.06}        \\ \hline
        RandAug~\cite{cubuk2020randaugment}  & 49.74 &  50.15 & \textbf{53.7}\\ \hline
        TrivialAug~\citep{muller2021trivialaugment}             &  47.42  &       47.98         & \textbf{51.22}\\ \hline
        PixMix~\citep{hendrycks2022pixmix}             & 58.57      & 58.38              & \textbf{59.60}\\ \hline
        \end{tabular}
        }
        
        \caption{Across different augmentation protocols, our proposed regularization provides non-trivial gains over standard training. Here, we show the accuracies for the challenging case of highest corruption severity.}
        \label{tab:aug}
    \end{subfigure}
    \hfill
    \begin{subfigure}[c]{0.38\textwidth}
        \centering
        \includegraphics[width=0.9\linewidth]{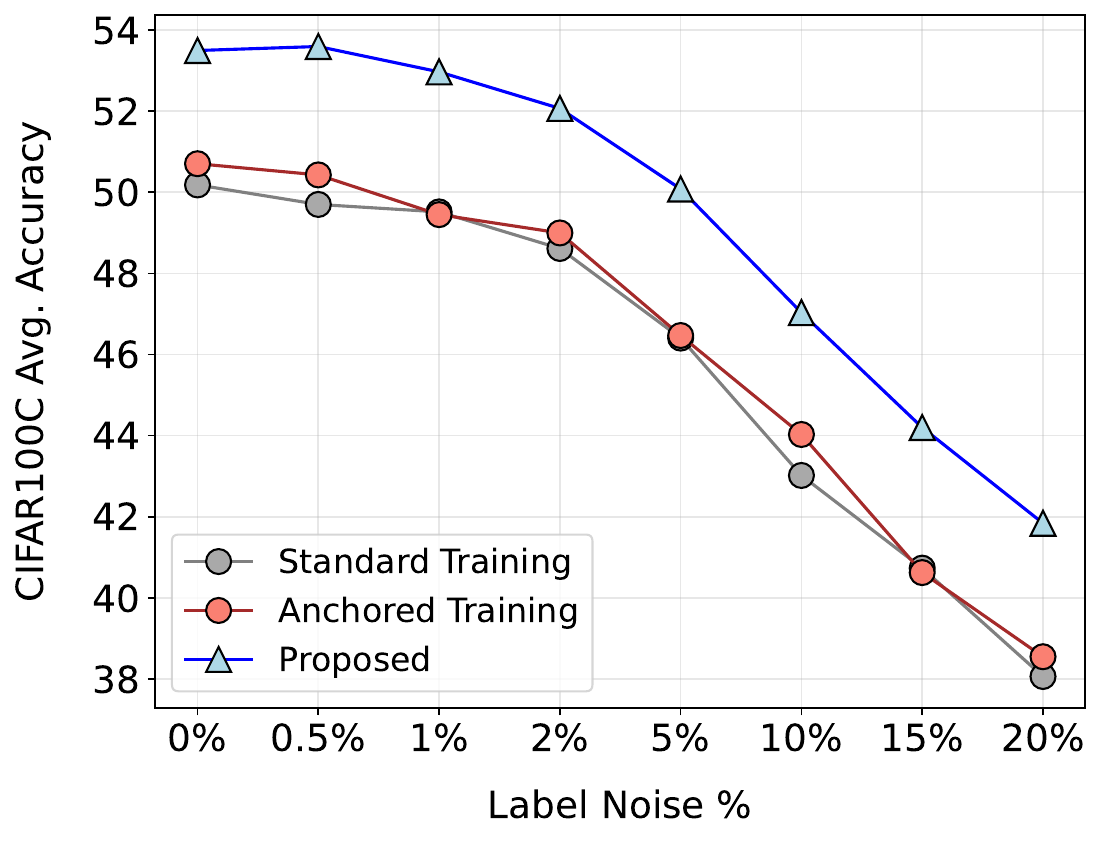} 
        \caption{Our approach demonstrates improved robustness to label noise in comparison to existing approaches.}
        \label{fig:ln}
    \end{subfigure}
    \caption{\textbf{Analysis of Anchored Models}. Using evaluations on the CIFAR100C OOD generalization of ResNet18 models trained on CIFAR100, we study the behavior of the proposed approach when combined with data augmentation protocols (left) and in presence of training label noise (right).}
    \label{fig:analysis}
\end{figure}

\subsection{Analysis}
\noindent\textbf{How does the accuracy landscape look like?} We hypothesize that, the improved generalization of anchoring stems from the training process itself, which inherently enables the model to find better solutions in the weight space. To validate this, we follow the analysis in~\citep{izmailov2018averaging}, where it was shown that that a well-generalizable solution is typically associated with a wider or flatter local optima in the loss/accuracy landscape. To this end, following the open-source implementation from~\cite{visualloss}, we obtained 2D heatmaps of accuracy evaluated on the CIFAR100C benchmark over different weight perturbations from the local minima inferring using diferent training strategies. \Cref{fig:landscape} visualizes the accuracy landscapes, where the $x$ and $y$ axes represent the co-ordinates that correspond to the different weight realizations. It can be observed that our approach produces wider and flatter optima in comparison to the baselines, thus explaining the generalization behavior.

\noindent\textbf{Can anchoring be combined with data augmentations?} Using synthetic data augmentations during training is a widely adopted method for improving generalization of vision models. In this study, we investigate if anchoring can be utilized alongside existing augmentation protocols, including state-of-the-art techniques like PixMix~\cite{hendrycks2022pixmix}), and if the observed generalization improvements persist. Table \ref{tab:aug} shows the CIFAR100C accuracies of models trained with different augmentation protocols. Note that, the architecture and the hyper-parameters of the augmentation protocols were fixed to be the same for a fair comparison. Remarkably, our approach consistently provides performance gains regardless of the augmentation protocols used, evidencing its utility as a generic training technique.  

\noindent\textbf{Does training label noise impact anchoring?} In practice, we construct the reference set $\mathcal{R} \subseteq \mathcal{D}$ for anchored training.  However, under label noise, a fraction (or all) noisy samples can be included in the reference set, and get used for obtaining relative representations. A natural question is if this will impact the anchored training; however, we remind that the tuple construction in anchoring does not use the target label of a reference, and the benefits of anchoring will persist even under label noise corruptions. We validate this using the following experiment: We randomly flip the labels of $l\%$ ($l = \{0.5, 1, 2, 5, 10, 15, 20\}$) of  training samples before training a ResNet18 model on CIFAR100, and evaluate the generalization performance on CIFAR100C. \Cref{fig:ln} illustrates that, with increasing levels of label noise, the anchored models do not demonstrate any additional challenges in handling label noise. In fact, it provides superior generalization ($\sim 4\%$ improvements at 20\% label noise) when compared to the standard and vanilla anchored training protocols. 

\section{Experiments}
\label{sec:experiments}
In this section, we empirically demonstrate the effectiveness of our proposed strategy in training models of varying scales (ResNets, Transformers) on datasets of different complexities (CIFAR10, CIFAR100, ImageNet). We systematically evaluate the generalization of these models under natural covariate shifts and synthetic corruptions. Additionally, we perform a comprehensive evaluation of model calibration, anomaly rejection, and robustness of task adapters in an effort to assess the  safety of anchored models.  

\noindent\textbf{Training Datasets.}
\underline{(i) CIFAR-10} and \underline{(ii) CIFAR-100}~\citep{krizhevsky2009learning} datasets contain $50,000$ training samples and $10,000$ test samples each of size $32\times32$ belonging to $10$ and $100$ classes, respectively; \underline{(iii) ImageNet-1K}~\citep{russakovsky2015imagenet} is a large-scale vision benchmark comprising $1.3$ million training images and $50,000$ validation images across 1000 diverse categories. 

\noindent\textbf{Architectures.}
We utilize a suite of vision transformer and CNN architectures with varying levels of structural and parameter complexity. Specifically for training with ImageNet, we consider SWINv2-T~($28.4$M params), SWINv2-S~($49.7$M), SWINv2-B~($87.8$M)~\citep{liu2021swinv2} and ViT-B-16~($86.6$M)~\citep{dosovitskiy2021an}.
For CIFAR100, we use ResNet-18~($11.7$M)~\citep{he2016deep} and WideResNet40-2~($2.2$M)~\citep{zagoruyko2016wide} architectures, and ResNet-18 for CIFAR10. We provide the training recipes adopted for our models in Section \ref{sec:training_protocols}.

\noindent\textbf{Choice of $\alpha$}. 
Through extensive empirical studies with multiple architectures, we found using the masking schedule hyper-parameter $\alpha=0.2$ (corresponds to every $5^{\text{th}}$ batch in an epoch), leads to stable convergence (closely match the top-$1$ validation accuracy of standard training) on ImageNet and $\alpha=0.25$ for CIFAR10/100. Note that, our approach performs reference masking for an entire batch as determined by $\alpha$. We have included our analysis on the impact of choice of $\alpha$ in Section \ref{sec:alpha}.

\subsection{Generalization to Covariate Shifts and Synthetic Corruptions}
\label{sec:OOD}
\noindent\textbf{OOD Datasets and Evaluation Metrics}. For models trained on CIFAR10, we evaluate generalization on CIFAR10C and CIFAR10$\bar{\text{C}}$. While the former contains $19$ different types of corruptions (e.g., noise, blur, weather, digital), CIFAR10$\bar{\text{C}}$ comprises $10$ types of synthetic noise, at 5 different severity levels respectively. Equivalently, for CIFAR100, we use the CIFAR100C and CIFAR100$\bar{\text{C}}$ benchmarks. For ImageNet-1K, we consider (i) {ImageNet-C}~\citep{hendrycks2018benchmarking} with $19$ natural image corruptions across $5$ severity levels, (ii) ImageNet-$\bar{\text{C}}$~\citep{mintun2021interaction} with $10$ noise corruptions across $5$ severity levels; (iii) ImageNet-R~\citep{hendrycks2021many} containing different renditions of $200$ classes from ImageNet; (iv) ImageNet-S~\citep{wang2019learning} comprising black and white sketch images from each class of ImageNet. 
We use the top@1 accuracy to evaluate generalization performance.

\begin{table*}[t]
\centering
\renewcommand{\arraystretch}{1.4}
\caption{\textbf{Generalization performance of CNNs trained on CIFAR10/100}. We report the ID test and the OOD (CIFAR10 -C/$\bar{\text{C}}$, CIFAR100 - C/$\bar{\text{C}}$) accuracies of standard and anchored CNNs to evaluate generalization ($\uparrow$). Note, we provide the difference ($\Delta$) between the proposed and the standard model in each case with \color[HTML]{B71BAA}{pink}.}
\label{tab:cifar}
\resizebox{\textwidth}{!}{
\begin{tabular}{c|c|c|c|ccccc|ccccc}
\toprule

 &  &  &  &  \multicolumn{5}{c|}{\textbf{CIFAR{10/100}-C Accuracy \%}} &\multicolumn{5}{c}{\textbf{CIFAR{10/100}-$\bar{\text{C}}$ Accuracy \%}} \\ 
\cline{5-14}
\textbf{Dataset} & \textbf{Model} & \textbf{Method}  & \textbf{ID Acc.} & \textbf{Sev. 1} & \textbf{Sev. 2} & \textbf{Sev. 3} & \textbf{Sev. 4} & \textbf{Sev. 5} & \textbf{Sev. 1} & \textbf{Sev. 2} & \textbf{Sev. 3} & \textbf{Sev. 4} & \textbf{Sev. 5} \\ 
\midrule
 &  & Standard  & $95.15$ & $89.44$ & $83.47$ & $77.91$ & $70.74$ & $58.72$ & $86.86$ & $81.97$ & $74.51$ & $65.94$ & $60.31$ \\ 

 & & Vanilla Anchoring & $94.92$ & $88.99$ & $84.28$ & $79.16$ & $72.09$ & $59.82$ & $87.04$ & $82.79$ & $75.00$ & $66.73$ & $61.52$ \\ 

& & Proposed & $95.72$ & $90.98$ & $87.15$ & $83.17$ & $77.81$ & $67.26$ & $89.24$ & $85.38$ & $78.34$ & $70.33$ & $65.43$ \\ 

\multirow{-4}{*}{CIFAR-10} & \multirow{-4}{*}{ResNet-18} &  \color[HTML]{B71BAA}$\Delta$ & \color[HTML]{B71BAA}$+0.57$&\color[HTML]{B71BAA}$+1.54$ &\color[HTML]{B71BAA}$+3.68$& \color[HTML]{B71BAA}$+5.26$ & \color[HTML]{B71BAA}$+7.07$ &\color[HTML]{B71BAA}$+8.54$ & \color[HTML]{B71BAA}$+2.38$ &\color[HTML]{B71BAA}$+3.41$&\color[HTML]{B71BAA}$+3.83$ & \color[HTML]{B71BAA}$+4.40$ & \color[HTML]{B71BAA}$+5.12$  \\

\midrule
 &  & Standard & $77.6$ & $65.56$ & $56.77$ & $51.25$ & $44.57$ & $34.13$ & $62.0$  & $54.08$  &$44.89$  & $36.55$  & $32.27$  \\ 

 &  & Vanilla Anchoring & $77.21$ & $65.67$ & $57.3$ & $52.02$ & $45.27$ & $34.79$ & $61.69$ & $54.17$  &  $44.98$& $36.90$  & $32.72$ \\ 

 &  & Proposed & $77.89$ & $67.0$ & $59.51$ & $54.88$ & $48.78$ & $38.66$ &   $64.47$&   $58.10$ &  $49.78$  &  $41.42$ &  $36.81$ \\ 

 & \multirow{-4}{*}{ResNet-18} & \color[HTML]{B71BAA}$\Delta$ & \color[HTML]{B71BAA}$+0.29$ & \color[HTML]{B71BAA}$+1.44$ & \color[HTML]{B71BAA}$+2.74$ & \color[HTML]{B71BAA}$+3.63$ & \color[HTML]{B71BAA}$+4.21$ & \color[HTML]{B71BAA}$+4.53$ & \color[HTML]{B71BAA}$+2.47$ & \color[HTML]{B71BAA}$+4.02$ & \color[HTML]{B71BAA}$+4.89$ & \color[HTML]{B71BAA}$+4.87$ & \color[HTML]{B71BAA}$+4.54$
 \\
 \cline{2-14}

 &  & Standard & $75.48$ & $62.26$ & $52.82$ & $46.85$ & $40.12$ & $30.05$ & $60.09$ & $52.89$ & $44.44$ & $35.78$ & $31.06$  \\ 

 & & Vanilla Anchoring & $76.67$ & $64.55$ & $55.47$ & $49.43$ & $42.84$ & $32.75$ & $61.59$ & $54.42$  & $45.50$  & $36.12$  & $31.11$  \\ 

&  & Proposed & $77.03$ & $66.0$ & $57.77$ & $52.33$ & $45.64$ & $35.52$ & $63.83$  & $57.76$ & $49.32$  & $40.26$ & $35.29$ \\ 
 \multirow{-8}{*}{CIFAR-100} & \multirow{-4}{*}{WRN 40-2} & \color[HTML]{B71BAA}$\Delta$ & \color[HTML]{B71BAA}$+1.55$ & \color[HTML]{B71BAA}$+3.74$ &\color[HTML]{B71BAA}$+4.95$ &\color[HTML]{B71BAA}$+5.48$ & \color[HTML]{B71BAA}$+5.52$ & \color[HTML]{B71BAA}$+5.47$ & \color[HTML]{B71BAA}$+3.74$ & \color[HTML]{B71BAA}$+4.87$ & \color[HTML]{B71BAA}$+4.88$ & \color[HTML]{B71BAA}$+4.48$ & \color[HTML]{B71BAA}$+4.23$
 \\ 

\midrule
\end{tabular}
}
\end{table*}

\begin{table*}[t]
\centering
\renewcommand{\arraystretch}{1.4}
\caption{\textbf{Generalization performance of different transformer architectures trained on ImageNet-1K}. We report the ID test and OOD (corruptions and covariate shifts) generalization performance of standard and anchored vision transformers using the top1 accuracy. For calibration performance, we report the mean and standard deviation of the Smoothed ECE ($\downarrow$) metric across all ImageNet OOD datasets. Note, we provide the difference ($\Delta$) between the proposed and the standard model in each case with \color[HTML]{B71BAA}{pink}.}
\label{tab:ood}
\resizebox{1\textwidth}{!}{
\begin{tabular}{c|ccc|ccc|ccc|ccc}
\toprule
& \multicolumn{3}{c|}{\textbf{SWINv2-T} ($28.4$M)} & \multicolumn{3}{c|}{\textbf{SWINv2-S} ($49.7$M)} & \multicolumn{3}{c|}{\textbf{VITb16} ($86.6$M)} & \multicolumn{3}{c}{\textbf{SWINv2-B} ($87.8$M)} \\
\cline{2-13}
\multirow{-2}{*}{\textbf{Dataset}} & \textbf{Standard} & \textbf{Proposed} & \color[HTML]{B71BAA}$\Delta$ & \textbf{Standard} & \textbf{Proposed} & \color[HTML]{B71BAA}$\Delta$& \textbf{Standard} & \textbf{Proposed} & \color[HTML]{B71BAA}$\Delta$&\textbf{Standard} & \textbf{Proposed}& \color[HTML]{B71BAA}$\Delta$\\
\midrule
{ImageNet (val)} & $82.07$ & $82.03$ &  \color[HTML]{B71BAA}$-0.04$&$83.71$ & $83.68$ & \color[HTML]{B71BAA}$-0.03$&$81.07$ & $80.76$ & \color[HTML]{B71BAA}$-0.31$&$84.11$ & $84.09$  & \color[HTML]{B71BAA}$-0.02$\\
\midrule
{ImageNet-R} & $40.84$ & $41.17$ &\color[HTML]{B71BAA}$+0.33$ &$45.17$ & $46.63$ & \color[HTML]{B71BAA}$+1.46$&$44.06$ & $46.39$ & \color[HTML]{B71BAA}$+2.33$&$45.7$ & $48.16$ & \color[HTML]{B71BAA}$+2.46$\\
\midrule
{ImageNet-S} & $27.08$ & $27.68$ & \color[HTML]{B71BAA}$+0.60$&$32.25$ & $33.3$ & \color[HTML]{B71BAA}$+1.05$&$29.4$ & $33.0$ & \color[HTML]{B71BAA}$+3.60$&$31.91$ & $33.34$ & \color[HTML]{B71BAA}$+1.43$\\
\midrule
{ImageNet-C (Sev. 1)} & $71.63$ & $72.13$ & \color[HTML]{B71BAA}$+0.50$&$74.48$ & $74.7$ & \color[HTML]{B71BAA}$+0.22$&$72.37$ & $72.52$ & \color[HTML]{B71BAA}$+0.15$& $74.45$ & $75.24$ & \color[HTML]{B71BAA}$+0.79$\\
{ImageNet-C (Sev. 2)} & $64.89$ & $65.71$ & \color[HTML]{B71BAA}$+0.82$&$68.8$ & $69.12$ & \color[HTML]{B71BAA}$+0.32$&$66.57$ & $67.38$ & \color[HTML]{B71BAA}$+0.81$&$68.55$ & $69.63$ & \color[HTML]{B71BAA}$+1.08$\\
{ImageNet-C (Sev. 3)} & $57.77$ & $59.21$ & \color[HTML]{B71BAA}$+1.44$&$62.84$ & $63.65$ & \color[HTML]{B71BAA}$+0.81$&$61.6$ & $62.87$ & \color[HTML]{B71BAA}$+1.27$&$62.34$ & $64.05$ & \color[HTML]{B71BAA}$+1.71$\\
{ImageNet-C (Sev. 4)} & $47.77$ & $50.01$ & \color[HTML]{B71BAA}$+2.24$&$54.32$ & $55.5$ & \color[HTML]{B71BAA}$+1.18$&$52.88$ & $55.13$ & \color[HTML]{B71BAA}$+2.25$&$53.66$ & $56.08$ & \color[HTML]{B71BAA}$+2.42$\\
{ImageNet-C (Sev. 5)} & $35.66$ & $38.58$ & \color[HTML]{B71BAA}$+2.92$&$42.85$ & $44.33$ & \color[HTML]{B71BAA}$+1.48$&$41.09$ & $44.52$ & \color[HTML]{B71BAA}$+3.43$&$41.87$ & $45.19$ & \color[HTML]{B71BAA}$+3.32$\\
\midrule
{ImageNet-$\bar{\text{C}}$ (Sev. 1)} & $71.37$ & $73.51$ & \color[HTML]{B71BAA}$+2.14$&$75.39$ & $76.59$ & \color[HTML]{B71BAA}$+1.20$&$72.75$ & $73.65$ & \color[HTML]{B71BAA}$+0.90$&$75.12$ & $77.1$& \color[HTML]{B71BAA}$+1.98$\\
{ImageNet-$\bar{\text{C}}$ (Sev. 2)} & $67.12$ & $70.45$ & \color[HTML]{B71BAA}$+3.33$&$72.26$ & $74.24$ & \color[HTML]{B71BAA}$+1.98$&$69.01$ & $70.91$ & \color[HTML]{B71BAA}$+1.90$&$72.15$ & $74.69$ &\color[HTML]{B71BAA}$+2.54$\\
{ImageNet-$\bar{\text{C}}$ (Sev. 3)} & $61.2$ & $65.77$ & \color[HTML]{B71BAA}$+4.57$&$67.14$ & $70.17$ &   \color[HTML]{B71BAA}$+3.03$ &$63.47$ & $66.87$ & \color[HTML]{B71BAA}$+3.39$&$67.16$ & $70.81$ &\color[HTML]{B71BAA}$+3.65$\\
{ImageNet-$\bar{\text{C}}$ (Sev. 4)} & $52.01$ & $57.31$ & \color[HTML]{B71BAA}$+5.30$&$58.73$ & $62.93$ & \color[HTML]{B71BAA}$+4.20$&$54.7$ & $59.29$ & \color[HTML]{B71BAA}$+4.59$&$58.66$ & $63.53$ &\color[HTML]{B71BAA}$+4.87$\\
{ImageNet-$\bar{\text{C}}$ (Sev. 5)} & $46.54$ & $51.76$ & \color[HTML]{B71BAA}$+5.22$&$53.7$ & $58.25$ & \color[HTML]{B71BAA}$+4.55$&$50.07$ & $54.94$ & \color[HTML]{B71BAA}$+4.86$&$53.75$ & $58.77$ &\color[HTML]{B71BAA}$+5.02$\\
\bottomrule
\end{tabular}
}
\end{table*}

\noindent\textbf{Results and Discussions}. First, in Table~\ref{tab:cifar}, we report the averaged accuracy over all corruptions for every severity level on the CIFAR10C/$\bar{\text{C}}$, CIFAR100C/$\bar{\text{C}}$ datasets, for the conv-nets trained on CIFAR10/100 respectively. We make a key finding that our proposed approach leads to significant gains in corruption accuracies across all severity levels over standard training ($1.54\% - 8.54\%)$ on an average. When compared to CIFAR10, the improvements of anchoring are apparent even at lower severity levels, for e.g., $+3.74$ improvement with WRN 40-2 at CIFAR100C severity level 1. 

Second, as shown in Table~\ref{tab:ood}, we investigated the efficacy of anchored transformers trained on the large-scale ImageNet-1K dataset in terms of OOD generalization. It can be observed that our proposed approach consistently yields improvements in corruption accuracies over standard training across all architectures. A striking observation is that network capacity plays a significant role in effectively leveraging the increased diversity produced by anchored training (we used the entire ImageNet-1K as the reference set). For example, as we move from SWINv2-T ($28.4$M) to SWINv2-B ($88$M), we observe increasingly larger performance gains over standard training. Importantly, our proposed strategy handles high noise severity better, achieving improvements of $2\% - 7\%$ at severity 5 for both Imagenet-C and $\bar{\text{C}}$. All these observations clearly evidence the importance of leveraging the diversity of $P_{(\mathrm{r}, \Delta)}$ for enhanced generalization. Finally, we observe from Tables~\ref{tab:cifar} and~\ref{tab:ood} that anchored training maintains competitive, and in a few cases, improved ID accuracies compared to standard training.

\subsection{Assessing Safety of Anchored Models} 
\label{sec:anomaly}
\textbf{Calibration and Anomaly Rejection}. While generalization is key to improve model utility, 
\begin{wraptable}{r}{0.6\textwidth}
    \centering
    \vspace{-5pt} 
    \renewcommand{\arraystretch}{1.0}
    \caption{\textbf{Anomaly rejection and calibration performance of transformers trained on ImageNet-1K}. We compare the anomaly rejection performance against standard training  using common vision OOD benchmarks (Textures, Places365, and iSUN datasets) and the more recent NINCO dataset. For evaluation, we consider the AUROC ($\uparrow$) metric. Moreover, we also provide Smoothed ECE scores ($\downarrow$) (mean, std) across different Imagenet corruption benchmarks. We highlight the best performing model in each case with \color[HTML]{B71BAA}{pink}.}
    \label{tab:anomaly}
    \resizebox{0.6\textwidth}{!}{
    \begin{tabular}{c|c|c|c|c}
    \toprule
     &   &  &   \\ 
    \textbf{Model} & \textbf{Method} & \textbf{Vision OOD} & \textbf{NINCO} &\textbf{Calibration} \\ 
    \midrule
     & Standard & $76.54$ & $77.46$ & $0.121\pm0.034$ \\ 
    \multirow{-2}{*}{SWINv2-T} & Proposed & \color[HTML]{B71BAA}$\mathbf{77.65}$ & \color[HTML]{B71BAA}$\mathbf{78.49}$ &  \color[HTML]{B71BAA}$\mathbf{0.117\pm0.027}$ \\ 
    \midrule
     & Standard & $77.13$ & $74.73$ & $0.126\pm0.039$\\ 
    \multirow{-2}{*}{SWINv2-S} & Proposed & $79.56$ & \color[HTML]{B71BAA}$\mathbf{78.47}$ & \color[HTML]{B71BAA}$\mathbf{0.119\pm0.041}$\\ 
    \midrule
     & Standard & \color[HTML]{B71BAA}$\mathbf{77.29}$ & $65.98$ & $0.109\pm0.037$\\ 
    \multirow{-2}{*}{VITb16} & Proposed & $76.88$ & \color[HTML]{B71BAA}$\mathbf{70.32}$ & \color[HTML]{B71BAA}$\mathbf{0.105\pm0.028}$ \\ 
    \midrule
     & Standard & $75.89$ & $72.13$ & $0.132\pm0.055$\\ 
    \multirow{-2}{*}{SWINv2-B} & Proposed & \color[HTML]{B71BAA}$\mathbf{78.91}$ & \color[HTML]{B71BAA}$\mathbf{74.53}$ & \color[HTML]{B71BAA}$\mathbf{0.124\pm0.051}$  \\ 
    \midrule
    \end{tabular}
    }
\end{wraptable}it must be ensured that the models are not over-confident on unknown inputs and produce well-calibrated prediction probabilities that match the likelihood of correctness. Hence, measuring calibration~\cite{guo2017calibration} is vital to understand how tempered the model predictions are under distribution shifts. On the other hand, when the inputs are semantically disconnected and do not share the same label space as the training data, we require the models to appropriately flag them as anomalies. To that end, we also conduct an extensive evaluation of model calibration under distribution shifts and anomaly rejection. For the former, we use the ImageNet-C/$\bar{\text{C}}$/R/S variants, and for the latter, we consider two benchmarks: (a) Vision OOD, comprising commonly used anomaly rejection datasets - \textit{iSUN}~\citep{xiao2010sun}, \textit{Textures}~\citep{DTD}, and \textit{Places365}~\citep{zhou2017places}; and (b) \textit{NINCO}~\citep{ninco}, a recent benchmark containing images with semantic overlap with ImageNet but with no class overlap. Following standard practice~\citep{liu2020energy}, we use the Smoothed ECE metric~\citep{blasiok2023smooth} to assess calibration. For anomaly rejection, we obtain the energy scores~\cite{liu2020energy} for both ID validation and OOD data, and report the AUROC metric. 

We report the anomaly rejection and calibration performance of of transformer models trained with ImageNet-1K in Table~\ref{tab:anomaly}. The results demonstrate notable improvements in anomaly rejection across architectures, highlighting the ability of our approach to better recognize residuals $\mathrm{x}_t-\bar{\mathrm{r}} = \bar{\mathrm{d}} \notin P_\Delta$ for an anamolous input sample $\mathrm{x}_t$ and a reference $\bar{\mathrm{r}}$ observed during training. This is evidenced by substantial gains on both vision OOD and the challenging NINCO anomaly detection benchmarks. For instance, ViTb16 trained with the proposed approach achieves a gain of $+4.34\% $ on AUROC over non-anchored variant on the NINCO benchmark. In addition, our approach produces consistently lower calibration errors irrespective of the choice of architecture, showcasing our ability to produce tempered predictions under OOD shifts.

\noindent{\textbf{Robustness to Task Adaptation}}. Evaluating model adaptation under task shifts~\citep{andreassen2021evolution} becomes important to shed light onto the quality and re-usability of features inferred in a backbone network. To that end we employ two evaluation protocols: Adaptation\texttt{(ID Eval)} and Adaptation (\texttt{OOD Eval}). In the former, we assume that the distribution of the dataset used for linear probing is the same as that of the test set. In the latter, we first train the linear probe with our anchored training approach using a probing dataset but evaluate the same with data drawn from a shifted wrt the probing dataset. Note, for both evaluation protocols, we fix the ViTb16 architecture as the Imagenet pre-trained feature extractor backbone. Note, we set $\alpha = 0.4$, a higher value than the original task model training as we observed stable convergence.

\begin{figure*}
    \centering
    \begin{subfigure}[c]{0.4\textwidth}
        \centering
        \renewcommand{\arraystretch}{1.4}
         \resizebox{1\textwidth}{!}{
        \begin{tabular}{c|cccc}
        \toprule
        & \multicolumn{3}{c}{\textbf{VITb16} ($86.6$M)} &  \\
        \cline{2-4}
        \multirow{-2}{*}{\textbf{Dataset}}                      & \textbf{Standard} & \textbf{Proposed} & \color[HTML]{B71BAA}$\Delta$ &  \\
        \cline{1-4}
        CIFAR-10      & $95.48$    & $96.29$    & \color[HTML]{B71BAA}$+0.81$  &  \\
        CIFAR-100     & $80.1$    & $82.78$    & \color[HTML]{B71BAA}$+2.68$  &  \\
        UCF101      & $75.55$    & $77.01$    & \color[HTML]{B71BAA}$+1.46$  &  \\
        Flowers102    & $94.68$    & $95.7$     & \color[HTML]{B71BAA}$+1.02$  &  \\
        StandfordCars & $58.54$    & $61.15$    & \color[HTML]{B71BAA}$+2.61$ & \\
        \bottomrule
        \end{tabular}}
         \caption{\textbf{LP-based adaptation for ViTb16 architecture pre-trained on Imagenet-1K on downstream tasks}. We measure the accuracy ($\uparrow$) of the adapted model using the validation split of the target dataset.}
    \end{subfigure}\hfill
    \begin{subfigure}[c]{0.58\textwidth}
        \centering
        \renewcommand{\arraystretch}{1.4}
        \label{tab:ood_lp}
        \resizebox{1.\textwidth}{!}{
        \begin{tabular}{c|ccc|ccc}
        \toprule
        \textbf{Evaluation}& \multicolumn{3}{c|}{Train Domain: \textbf{Real} } & \multicolumn{3}{c}{Train Domain: \textbf{Sketch}}  \\
        \cline{2-7}
        {\textbf{Domain}} & \textbf{Standard} & \textbf{Proposed} & \color[HTML]{B71BAA}$\Delta$ & \textbf{Standard} & \textbf{Proposed} & \color[HTML]{B71BAA}$\Delta$\\
        \midrule
        {Real} & $-$ & $-$ &  $-$&$41.35$ & $44.81$ & \color[HTML]{B71BAA}$+3.46$\\
        {Sketch} & $25.85$ & $28.02$ &  \color[HTML]{B71BAA}$+2.17$&$-$ & $-$ & $-$\\
        {Clipart} & $37.38$ & $38.98$ &  \color[HTML]{B71BAA}$+1.6$& $35.4$ & $37.76$ & \color[HTML]{B71BAA}$+2.36$\\
        {Painting} & $46.3$ & $46.97$ &  \color[HTML]{B71BAA}$+0.67$&$31.42$ & $32.7$ & \color[HTML]{B71BAA}$+1.28$\\
        \bottomrule
        \end{tabular}
        }
         \caption{\textbf{OOD Evaluation of LP Adaptation}. Using the ViTb16 backbone we train two LPs for the \textit{Real} and \textit{Sketch} domains from the Domainnet dataset respectively. We then assess their zero-shot accuracies on three held-out test domains. Our findings show that the proposed approach consistently outperforms the non-anchored baselines.}
    \end{subfigure}
    \caption{Assessing anchored and standard pre-trained ImageNet backbones on robustness to task shifts.}
    \label{tab:adaptation_id_ood}
\end{figure*}

\noindent \textbf{Adaptation (\texttt{ID Eval})}: We consider the following target datasets:
 (i) {CIFAR-10}~\citep{krizhevsky2009learning}  ;
 (ii) {CIFAR-100}~\citep{krizhevsky2014cifar} ;
 (iii) {UCF101}~\citep{ucf101};
 (iv) {Flowers102}~\citep{flowers102};
 (v) {StanfordCars}~\citep{standordcars}. The results in Figure \ref{tab:adaptation_id_ood}(a) demonstrate that the proposed approach achieves substantial performance gains over the baseline ($0.81\%$ - $2.68\%$). These findings suggest that the reference masking regularizer yields feature representations that are transferable even under complex task shifts. 

\noindent\textbf{Adaptation (\texttt{OOD Eval})}: For training linear probes, we use the DomainNet~\citep{domainnet}, comprising of images from $345$ categories across six diverse domains. Specifically, we pick four domains, namely \textit{real}, \textit{sketch}, \textit{clipart}, and \textit{painting} and train probes on (i) images from the \textit{real} domain, and (ii) images from the \textit{sketch} domain respectively. We then evaluate the LPs on the remaining three held-out domains.  As Figure~\ref{tab:adaptation_id_ood}(b) illustrates, our proposed reference masking continues to substantially outperform standard training baseline on all held-out domains under both configurations. We attribute this behavior to our approach being able to effectively leverage the diversity in the reference-residual space to produce robust and better generalizable features supporting transferability. 

\section{Related Work}
\noindent\textbf{Anchoring in Predictive Models}. Our work is based on the principle of anchoring first introduced in~\citep{thiagarajan2022single} where it was used to achieve stochastic data centering for epistemic uncertainty estimation. Since then, the  anchoring has been extended to a variety of use-cases and applications. For e.g,  Netanyahu \textit{et al.}~\citep{netanyahu2023learning} utilized anchoring for extrapolating to unseen data regimes~\citep{netanyahu2023learning} in regression settings and Trivedi \textit{et al.}~\citep{Trivedi23_GDUQ} employed the same for graph neural network calibration. In contrast, our paper is the first to explore and facilitate the utility of anchoring as a viable training protocol for large scale vision models.

\noindent\textbf{Data Augmentations}. Augmentation strategies enforce models to be robust under different pixel-space manipulations improving generalization. For e.g., strategies such as Augmix~\cite{hendrycks2020augmix} or random convolutions (RandConv)~\cite{xu2021robust} are known to improve generalization. Recent advancements in the field include strategies such as PixMix~\citep{hendrycks2022pixmix}, which utilizes an external dataset with complex image patterns to augment the training data, and ALT~\cite{gokhale2023improving}, which learns adversarially robust augmentations. While the idea of enforcing prediction consistency in anchoring might appear similar to training with synthetic data augmentations, we emphasize that anchoring does not alter the data (e.g., with perturbations or geometric transformations) but only creates relative representations for each sample with respect to different reference choices. Furthermore, it can be combined with data augmentations to achieve further gains in generalization (Table \ref{tab:aug}).  

\noindent\textbf{Model Safety}. As models are being increasingly adopted in a variety of sensitive applications~\citep{davenport2019potential, bogdoll2022anomaly}, safe model deployment has become critical~\citep{barrett2023identifying, hendrycks2021unsolved}. In this context, generalization to data beyond the training distribution~\citep{yang2021generalized, hendrycks2018benchmarking}, ability to accurately detect anomalies in the input data~\cite{hendrycks17baseline,liu2020energy, hendrycks2018deep} as well producing calibrated prediction probabilities~\cite{guo2017calibration, anirudh2023out} are all important facets of safety evaluation. Hendrycks \textit{et al.}~\citep{hendrycks2022pixmix} argued that most existing training strategies compromise for one safety objective to satisfy another objective, thus limiting their real-world utility. We find from our experiments that anchoring jointly produces better generalization, calibration and anomaly rejection properties, which makes it a promising choice for practical deployment.  

\section{Conclusion}
Through this work, we showed that anchoring leads to significant performance gains in generalization and other safety metrics, including calibration, anomaly rejection, and task adaptation, across varying dataset sizes (CIFAR-10 to ImageNet) and model architectures (Conv-Nets to Transformers). Notably, when the training recipe includes high-capacity architectures or advanced mechanisms, our method yields even greater performance gains over the base models. Our observations suggest that anchored training with larger reference sets requires reference masking regularization to control the risk of learning undesirable shortcuts while making predictions. However, we realize that state-of-the-art results in OOD generalization are often obtained using model souping~\citep{souping} or by fine-tuning large scale pre-trained models~\citep{Goyal_2023_CVPR}. Hence, we believe it will be valuable to integrate anchoring into these approaches. While we have not theoretically characterized the generalization of anchored models, our hypothesis is rooted in existing theory and our empirical results provide evidence for the hypothesis. However, developing a theoretical understanding of anchored models is crucial and forms an important future direction.

\section*{Acknowledgements}
This work was performed under the auspices of the U.S. Department of Energy by the Lawrence Livermore National Laboratory under Contract No. DE-AC52-07NA27344, Lawrence Livermore National Security, LLC. and was supported by the LLNL-LDRD Program under Project No. 22-ERD-006. LLNL-CONF-864979

\vskip 0.2in
\bibliography{refs}

\begin{thebibliography}{47}
\providecommand{\natexlab}[1]{#1}
\providecommand{\url}[1]{\texttt{#1}}
\expandafter\ifx\csname urlstyle\endcsname\relax
  \providecommand{\doi}[1]{doi: #1}\else
  \providecommand{\doi}{doi: \begingroup \urlstyle{rm}\Url}\fi

\bibitem[Andreassen et~al.(2021)Andreassen, Bahri, Neyshabur, and Roelofs]{andreassen2021evolution}
Anders Andreassen, Yasaman Bahri, Behnam Neyshabur, and Rebecca Roelofs.
\newblock The evolution of out-of-distribution robustness throughout fine-tuning.
\newblock \emph{arXiv preprint arXiv:2106.15831}, 2021.

\bibitem[Anirudh and Thiagarajan(2022)]{anirudh2022out}
Rushil Anirudh and Jayaraman~J Thiagarajan.
\newblock Out of distribution detection via neural network anchoring.
\newblock In \emph{Asian Conference on Machine Learning (ACML)}. PMLR, 2022.

\bibitem[Anirudh and Thiagarajan(2023)]{anirudh2023out}
Rushil Anirudh and Jayaraman~J Thiagarajan.
\newblock Out of distribution detection via neural network anchoring.
\newblock In \emph{Asian Conference on Machine Learning}, pages 32--47. PMLR, 2023.

\bibitem[Barrett et~al.(2023)Barrett, Boyd, Bursztein, Carlini, Chen, Choi, Chowdhury, Christodorescu, Datta, Feizi, et~al.]{barrett2023identifying}
Clark Barrett, Brad Boyd, Elie Bursztein, Nicholas Carlini, Brad Chen, Jihye Choi, Amrita~Roy Chowdhury, Mihai Christodorescu, Anupam Datta, Soheil Feizi, et~al.
\newblock Identifying and mitigating the security risks of generative ai.
\newblock \emph{Foundations and Trends{\textregistered} in Privacy and Security}, 6\penalty0 (1):\penalty0 1--52, 2023.

\bibitem[Bitterwolf et~al.(2023)Bitterwolf, Mueller, and Hein]{ninco}
Julian Bitterwolf, Maximilian Mueller, and Matthias Hein.
\newblock In or out? fixing imagenet out-of-distribution detection evaluation.
\newblock In \emph{ICML}, 2023.
\newblock URL \url{https://proceedings.mlr.press/v202/bitterwolf23a.html}.

\bibitem[B{\l}asiok and Nakkiran(2023)]{blasiok2023smooth}
Jaros{\l}aw B{\l}asiok and Preetum Nakkiran.
\newblock Smooth ece: Principled reliability diagrams via kernel smoothing.
\newblock \emph{arXiv preprint arXiv:2309.12236}, 2023.

\bibitem[Bogdoll et~al.(2022)Bogdoll, Nitsche, and Z{\"o}llner]{bogdoll2022anomaly}
Daniel Bogdoll, Maximilian Nitsche, and J~Marius Z{\"o}llner.
\newblock Anomaly detection in autonomous driving: A survey.
\newblock In \emph{Proceedings of the IEEE/CVF conference on computer vision and pattern recognition}, pages 4488--4499, 2022.

\bibitem[Cimpoi et~al.(2014)Cimpoi, Maji, Kokkinos, Mohamed, and Vedaldi]{DTD}
Mircea Cimpoi, Subhransu Maji, Iasonas Kokkinos, Sammy Mohamed, and Andrea Vedaldi.
\newblock Describing textures in the wild.
\newblock In \emph{Proceedings of the IEEE conference on computer vision and pattern recognition}, pages 3606--3613, 2014.

\bibitem[Cubuk et~al.(2020)Cubuk, Zoph, Shlens, and Le]{cubuk2020randaugment}
Ekin~D Cubuk, Barret Zoph, Jonathon Shlens, and Quoc~V Le.
\newblock Randaugment: Practical automated data augmentation with a reduced search space.
\newblock In \emph{Proceedings of the IEEE/CVF conference on computer vision and pattern recognition workshops}, pages 702--703, 2020.

\bibitem[Davenport and Kalakota(2019)]{davenport2019potential}
Thomas Davenport and Ravi Kalakota.
\newblock The potential for artificial intelligence in healthcare.
\newblock \emph{Future healthcare journal}, 6\penalty0 (2):\penalty0 94, 2019.

\bibitem[Dosovitskiy et~al.(2021)Dosovitskiy, Beyer, Kolesnikov, Weissenborn, Zhai, Unterthiner, Dehghani, Minderer, Heigold, Gelly, Uszkoreit, and Houlsby]{dosovitskiy2021an}
Alexey Dosovitskiy, Lucas Beyer, Alexander Kolesnikov, Dirk Weissenborn, Xiaohua Zhai, Thomas Unterthiner, Mostafa Dehghani, Matthias Minderer, Georg Heigold, Sylvain Gelly, Jakob Uszkoreit, and Neil Houlsby.
\newblock An image is worth 16x16 words: Transformers for image recognition at scale.
\newblock In \emph{International Conference on Learning Representations}, 2021.
\newblock URL \url{https://openreview.net/forum?id=YicbFdNTTy}.

\bibitem[Garipov et~al.(2018)Garipov, Izmailov, Podoprikhin, Vetrov, and Wilson]{garipov2018loss}
Timur Garipov, Pavel Izmailov, Dmitrii Podoprikhin, Dmitry~P Vetrov, and Andrew~G Wilson.
\newblock Loss surfaces, mode connectivity, and fast ensembling of dnns.
\newblock \emph{Advances in neural information processing systems}, 31, 2018.

\bibitem[Gokhale et~al.(2023)Gokhale, Anirudh, Thiagarajan, Kailkhura, Baral, and Yang]{gokhale2023improving}
Tejas Gokhale, Rushil Anirudh, Jayaraman~J Thiagarajan, Bhavya Kailkhura, Chitta Baral, and Yezhou Yang.
\newblock Improving diversity with adversarially learned transformations for domain generalization.
\newblock In \emph{Proceedings of the IEEE/CVF Winter Conference on Applications of Computer Vision}, pages 434--443, 2023.

\bibitem[Goyal et~al.(2023)Goyal, Kumar, Garg, Kolter, and Raghunathan]{Goyal_2023_CVPR}
Sachin Goyal, Ananya Kumar, Sankalp Garg, Zico Kolter, and Aditi Raghunathan.
\newblock Finetune like you pretrain: Improved finetuning of zero-shot vision models.
\newblock In \emph{Proceedings of the IEEE/CVF Conference on Computer Vision and Pattern Recognition (CVPR)}, pages 19338--19347, June 2023.

\bibitem[Guo et~al.(2017)Guo, Pleiss, Sun, and Weinberger]{guo2017calibration}
Chuan Guo, Geoff Pleiss, Yu~Sun, and Kilian~Q Weinberger.
\newblock On calibration of modern neural networks.
\newblock In \emph{International Conference on Machine Learning}, pages 1321--1330. PMLR, 2017.

\bibitem[He et~al.(2016)He, Zhang, Ren, and Sun]{he2016deep}
Kaiming He, Xiangyu Zhang, Shaoqing Ren, and Jian Sun.
\newblock Deep residual learning for image recognition.
\newblock In \emph{Proceedings of the IEEE conference on computer vision and pattern recognition}, pages 770--778, 2016.

\bibitem[Hendrycks and Dietterich(2019)]{hendrycks2018benchmarking}
Dan Hendrycks and Thomas Dietterich.
\newblock Benchmarking neural network robustness to common corruptions and perturbations.
\newblock In \emph{International Conference on Learning Representations}, 2019.
\newblock URL \url{https://openreview.net/forum?id=HJz6tiCqYm}.

\bibitem[Hendrycks and Gimpel(2017)]{hendrycks17baseline}
Dan Hendrycks and Kevin Gimpel.
\newblock A baseline for detecting misclassified and out-of-distribution examples in neural networks.
\newblock \emph{Proceedings of International Conference on Learning Representations}, 2017.

\bibitem[Hendrycks et~al.(2018)Hendrycks, Mazeika, and Dietterich]{hendrycks2018deep}
Dan Hendrycks, Mantas Mazeika, and Thomas Dietterich.
\newblock Deep anomaly detection with outlier exposure.
\newblock In \emph{International Conference on Learning Representations}, 2018.

\bibitem[Hendrycks et~al.(2020)Hendrycks, Mu, Cubuk, Zoph, Gilmer, and Lakshminarayanan]{hendrycks2020augmix}
Dan Hendrycks, Norman Mu, Ekin~D. Cubuk, Barret Zoph, Justin Gilmer, and Balaji Lakshminarayanan.
\newblock {AugMix}: A simple data processing method to improve robustness and uncertainty.
\newblock \emph{Proceedings of the International Conference on Learning Representations (ICLR)}, 2020.

\bibitem[Hendrycks et~al.(2021{\natexlab{a}})Hendrycks, Basart, Mu, Kadavath, Wang, Dorundo, Desai, Zhu, Parajuli, Guo, Song, Steinhardt, and Gilmer]{hendrycks2021many}
Dan Hendrycks, Steven Basart, Norman Mu, Saurav Kadavath, Frank Wang, Evan Dorundo, Rahul Desai, Tyler Zhu, Samyak Parajuli, Mike Guo, Dawn Song, Jacob Steinhardt, and Justin Gilmer.
\newblock The many faces of robustness: A critical analysis of out-of-distribution generalization.
\newblock \emph{ICCV}, 2021{\natexlab{a}}.

\bibitem[Hendrycks et~al.(2021{\natexlab{b}})Hendrycks, Carlini, Schulman, and Steinhardt]{hendrycks2021unsolved}
Dan Hendrycks, Nicholas Carlini, John Schulman, and Jacob Steinhardt.
\newblock Unsolved problems in ml safety.
\newblock \emph{arXiv preprint arXiv:2109.13916}, 2021{\natexlab{b}}.

\bibitem[Hendrycks et~al.(2022)Hendrycks, Zou, Mazeika, Tang, Li, Song, and Steinhardt]{hendrycks2022pixmix}
Dan Hendrycks, Andy Zou, Mantas Mazeika, Leonard Tang, Bo~Li, Dawn Song, and Jacob Steinhardt.
\newblock Pixmix: Dreamlike pictures comprehensively improve safety measures.
\newblock In \emph{Proceedings of the IEEE/CVF Conference on Computer Vision and Pattern Recognition}, pages 16783--16792, 2022.

\bibitem[Izmailov et~al.(2018)Izmailov, Podoprikhin, Garipov, Vetrov, and Wilson]{izmailov2018averaging}
Pavel Izmailov, Dmitrii Podoprikhin, Timur Garipov, Dmitry Vetrov, and Andrew~Gordon Wilson.
\newblock Averaging weights leads to wider optima and better generalization.
\newblock \emph{arXiv preprint arXiv:1803.05407}, 2018.

\bibitem[Jacot et~al.(2018)Jacot, Gabriel, and Hongler]{jacot2018neural}
Arthur Jacot, Franck Gabriel, and Cl{\'e}ment Hongler.
\newblock Neural tangent kernel: Convergence and generalization in neural networks.
\newblock \emph{Advances in neural information processing systems}, 31, 2018.

\bibitem[Krause et~al.(2013)Krause, Stark, Deng, and Fei-Fei]{standordcars}
Jonathan Krause, Michael Stark, Jia Deng, and Li~Fei-Fei.
\newblock 3d object representations for fine-grained categorization.
\newblock In \emph{4th International IEEE Workshop on 3D Representation and Recognition (3dRR-13)}, Sydney, Australia, 2013.

\bibitem[Krizhevsky et~al.(2009)Krizhevsky, Hinton, et~al.]{krizhevsky2009learning}
Alex Krizhevsky, Geoffrey Hinton, et~al.
\newblock Learning multiple layers of features from tiny images.
\newblock \emph{Masters Thesis}, 2009.

\bibitem[Krizhevsky et~al.(2014)Krizhevsky, Nair, and Hinton]{krizhevsky2014cifar}
Alex Krizhevsky, Vinod Nair, and Geoffrey Hinton.
\newblock The cifar-10 dataset.
\newblock \emph{online: http://www. cs. toronto. edu/kriz/cifar. html}, 55:\penalty0 5, 2014.

\bibitem[Li et~al.(2018)Li, Xu, Taylor, Studer, and Goldstein]{visualloss}
Hao Li, Zheng Xu, Gavin Taylor, Christoph Studer, and Tom Goldstein.
\newblock Visualizing the loss landscape of neural nets.
\newblock In \emph{Neural Information Processing Systems}, 2018.

\bibitem[Liu et~al.(2020)Liu, Wang, Owens, and Li]{liu2020energy}
Weitang Liu, Xiaoyun Wang, John Owens, and Yixuan Li.
\newblock Energy-based out-of-distribution detection.
\newblock \emph{Advances in Neural Information Processing Systems}, 2020.

\bibitem[Liu et~al.(2022)Liu, Hu, Lin, Yao, Xie, Wei, Ning, Cao, Zhang, Dong, Wei, and Guo]{liu2021swinv2}
Ze~Liu, Han Hu, Yutong Lin, Zhuliang Yao, Zhenda Xie, Yixuan Wei, Jia Ning, Yue Cao, Zheng Zhang, Li~Dong, Furu Wei, and Baining Guo.
\newblock Swin transformer v2: Scaling up capacity and resolution.
\newblock In \emph{International Conference on Computer Vision and Pattern Recognition (CVPR)}, 2022.

\bibitem[Mintun et~al.(2021)Mintun, Kirillov, and Xie]{mintun2021interaction}
Eric Mintun, Alexander Kirillov, and Saining Xie.
\newblock On interaction between augmentations and corruptions in natural corruption robustness.
\newblock \emph{Advances in Neural Information Processing Systems}, 34:\penalty0 3571--3583, 2021.

\bibitem[M{\"u}ller and Hutter(2021)]{muller2021trivialaugment}
Samuel~G M{\"u}ller and Frank Hutter.
\newblock Trivialaugment: Tuning-free yet state-of-the-art data augmentation.
\newblock In \emph{Proceedings of the IEEE/CVF international conference on computer vision}, pages 774--782, 2021.

\bibitem[Netanyahu et~al.(2023)Netanyahu, Gupta, Simchowitz, Zhang, and Agrawal]{netanyahu2023learning}
Aviv Netanyahu, Abhishek Gupta, Max Simchowitz, Kaiqing Zhang, and Pulkit Agrawal.
\newblock Learning to extrapolate: A transductive approach.
\newblock In \emph{The Eleventh International Conference on Learning Representations}, 2023.
\newblock URL \url{https://openreview.net/forum?id=lid14UkLPd4}.

\bibitem[Nilsback and Zisserman(2008)]{flowers102}
Maria-Elena Nilsback and Andrew Zisserman.
\newblock Automated flower classification over a large number of classes.
\newblock In \emph{2008 Sixth Indian Conference on Computer Vision, Graphics \& Image Processing}, pages 722--729. IEEE, 2008.

\bibitem[Peng et~al.(2019)Peng, Bai, Xia, Huang, Saenko, and Wang]{domainnet}
Xingchao Peng, Qinxun Bai, Xide Xia, Zijun Huang, Kate Saenko, and Bo~Wang.
\newblock Moment matching for multi-source domain adaptation.
\newblock In \emph{Proceedings of the IEEE/CVF international conference on computer vision}, pages 1406--1415, 2019.

\bibitem[Russakovsky et~al.(2015)Russakovsky, Deng, Su, Krause, Satheesh, Ma, Huang, Karpathy, Khosla, Bernstein, et~al.]{russakovsky2015imagenet}
Olga Russakovsky, Jia Deng, Hao Su, Jonathan Krause, Sanjeev Satheesh, Sean Ma, Zhiheng Huang, Andrej Karpathy, Aditya Khosla, Michael Bernstein, et~al.
\newblock Imagenet large scale visual recognition challenge.
\newblock \emph{International journal of computer vision}, 115\penalty0 (3):\penalty0 211--252, 2015.

\bibitem[Soomro et~al.(2012)Soomro, Zamir, and Shah]{ucf101}
Khurram Soomro, Amir~Roshan Zamir, and Mubarak Shah.
\newblock Ucf101: A dataset of 101 human actions classes from videos in the wild.
\newblock \emph{arXiv preprint arXiv:1212.0402}, 2012.

\bibitem[Thiagarajan et~al.(2022)Thiagarajan, Anirudh, Narayanaswamy, and timo Bremer]{thiagarajan2022single}
Jayaraman~J. Thiagarajan, Rushil Anirudh, Vivek Narayanaswamy, and Peer timo Bremer.
\newblock Single model uncertainty estimation via stochastic data centering.
\newblock In Alice~H. Oh, Alekh Agarwal, Danielle Belgrave, and Kyunghyun Cho, editors, \emph{Advances in Neural Information Processing Systems}, 2022.
\newblock URL \url{https://openreview.net/forum?id=j0J9upqN5va}.

\bibitem[Trivedi et~al.(2023)Trivedi, Heimann, Anirudh, Koutra, and Thiagarajan]{Trivedi23_GDUQ}
Puja Trivedi, Mark Heimann, Rushil Anirudh, Danai Koutra, and Jayaraman~J. Thiagarajan.
\newblock Estimating epistemic uncertainty of graph neural networks.
\newblock In \emph{Data Centric Machine Learning Workshop @ ICML}, 2023.

\bibitem[Wang et~al.(2019)Wang, Ge, Lipton, and Xing]{wang2019learning}
Haohan Wang, Songwei Ge, Zachary Lipton, and Eric~P Xing.
\newblock Learning robust global representations by penalizing local predictive power.
\newblock In \emph{Advances in Neural Information Processing Systems}, pages 10506--10518, 2019.

\bibitem[Wortsman et~al.(2022)Wortsman, Ilharco, Gadre, Roelofs, Gontijo-Lopes, Morcos, Namkoong, Farhadi, Carmon, Kornblith, and Schmidt]{souping}
Mitchell Wortsman, Gabriel Ilharco, Samir~Ya Gadre, Rebecca Roelofs, Raphael Gontijo-Lopes, Ari~S Morcos, Hongseok Namkoong, Ali Farhadi, Yair Carmon, Simon Kornblith, and Ludwig Schmidt.
\newblock Model soups: averaging weights of multiple fine-tuned models improves accuracy without increasing inference time.
\newblock In Kamalika Chaudhuri, Stefanie Jegelka, Le~Song, Csaba Szepesvari, Gang Niu, and Sivan Sabato, editors, \emph{Proceedings of the 39th International Conference on Machine Learning}, volume 162 of \emph{Proceedings of Machine Learning Research}, pages 23965--23998. PMLR, 17--23 Jul 2022.
\newblock URL \url{https://proceedings.mlr.press/v162/wortsman22a.html}.

\bibitem[Xiao et~al.(2010)Xiao, Hays, Ehinger, Oliva, and Torralba]{xiao2010sun}
Jianxiong Xiao, James Hays, Krista~A Ehinger, Aude Oliva, and Antonio Torralba.
\newblock Sun database: Large-scale scene recognition from abbey to zoo.
\newblock In \emph{2010 IEEE computer society conference on computer vision and pattern recognition}, pages 3485--3492. IEEE, 2010.

\bibitem[Xu et~al.(2021)Xu, Liu, Yang, Raffel, and Niethammer]{xu2021robust}
Zhenlin Xu, Deyi Liu, Junlin Yang, Colin Raffel, and Marc Niethammer.
\newblock Robust and generalizable visual representation learning via random convolutions.
\newblock In \emph{International Conference on Learning Representations}, 2021.

\bibitem[Yang et~al.(2021)Yang, Zhou, Li, and Liu]{yang2021generalized}
Jingkang Yang, Kaiyang Zhou, Yixuan Li, and Ziwei Liu.
\newblock Generalized out-of-distribution detection: A survey.
\newblock \emph{arXiv preprint arXiv:2110.11334}, 2021.

\bibitem[Zagoruyko and Komodakis(2016)]{zagoruyko2016wide}
Sergey Zagoruyko and Nikos Komodakis.
\newblock Wide residual networks.
\newblock In \emph{British Machine Vision Conference 2016}. British Machine Vision Association, 2016.

\bibitem[Zhou et~al.(2017)Zhou, Lapedriza, Khosla, Oliva, and Torralba]{zhou2017places}
Bolei Zhou, Agata Lapedriza, Aditya Khosla, Aude Oliva, and Antonio Torralba.
\newblock Places: A 10 million image database for scene recognition.
\newblock \emph{IEEE transactions on pattern analysis and machine intelligence}, 40\penalty0 (6):\penalty0 1452--1464, 2017.

\end{thebibliography}

\newpage

\appendix

\section{How does the choice of $\alpha$ impact training?}
\label{sec:alpha}
The parameter $\alpha$ controls the frequency of the regularization applied to anchored training. Under the assumptions of operating with wide reference sets, through Table~\ref{fig:alpha} we note that moderate to small values of $\alpha$ enable better regularization of anchored training. Notably, setting $\alpha=0.25$ i.e. masking references for one in four samples, yields impressive gains in ID and OOD performance. Conversely, over-regularizing by setting $\alpha$ to a large value (e.g $1.0$) entails masking every reference, unsurprisingly results in models that generalize poorly, as they are tasked with learning solely from residuals. 

\begin{table}[ht]
    \centering
    \renewcommand{\arraystretch}{1.2}
       \caption{\textbf{Impact of $\alpha$ on anchored training}. As we gradually increase $\alpha$, there is a risk of over-regularization which can lead to severe underfits. Note, we consider $\mathcal{R} = \mathcal{D}$ in this study.  }
    \setlength{\arrayrulewidth}{0.1mm}
    \setlength{\tabcolsep}{2.0pt}
    \begin{tabular}{|c|c|c|c|c|c|}
        \hline
        \rowcolor[HTML]{EFEFEF}
        \textbf{$\alpha \rightarrow$} & 0.0 & 0.25 & 0.5 & 0.75 & 1.0 \\ \hline
       ID Test Acc. \% & 77.21 & \textbf{77.89 }& 76.97 & 75.4 & 57.90\\
         \hline
       OOD Acc. \% & 51.01 & \textbf{53.77} & 52.61 & 52.30 & 35.40 \\
       \hline
    \end{tabular}
    \label{fig:alpha}
    
\end{table}

\begin{table*}[h]
\centering
\renewcommand{\arraystretch}{1.5}
\caption{\textbf{Protocols adopted for training anchored models across different datasets and architectures.} While we adopt standard training recipes for training our models, we note that anchoring can serve as a generic wrapper that can be applied on top of any other existing recipe.
}
\label{tab:recipe}
\resizebox{\textwidth}{!}{
\begin{tabular}{c|c|c|cc|c}
\toprule
 &  & & \multicolumn{2}{c|}{\textbf{Number of Epochs}} & \\ 
\cline{4-5}
\textbf{Model} & \textbf{Dataset} & \textbf{Training Recipes} & \textbf{Non-Anchored} & \textbf{Anchored} & \textbf{Optimizer}\\ 
\midrule
ResNet-18, WRN-40-2  & CIFAR-10/100                         & \texttt{Horizontal \& Vertical Flips}                                                                                                                                                     & 200                      & 200                 & SGD with Multi-Step           \\

\midrule
SWINv2-T, SWINv2-S, SWINv2-B  & ImageNet   & \texttt{Mixup}, \texttt{CutMix}, \texttt{AutoAugment}, \texttt{Random Erase}, \texttt{AugMix}, \texttt{Label Smoothing}                                                  & 300                     & 330                 & AdamW with Cosine Annealing         \\
\midrule
VITb16  & ImageNet                         & \texttt{Mixup}, \texttt{CutMix}, \texttt{AutoAugment}, \texttt{AugMix}, \texttt{Label Smoothing}                                                                         & 300                     & 330                 & AdamW with Cosine Annealing   \\
\midrule

\end{tabular}}
\end{table*}

\section{Additional Details on Training Protocols} 
\label{sec:training_protocols}
Table \ref{tab:recipe} outlines the recipes (augmentations, epochs, optimizers) leveraged for model training. Note that, the other hyper-parameters can be found in ~\citep{anirudh2022out} for CIFAR10/100 and \url{https://pytorch.org/vision/stable/models.html} for ImageNet. We emphasize that, anchoring can be used as a generic model training wrapper, allows integration with any data augmentation or training strategy, and is not restricted to the recipes considered.

\subsection{Expanded ImageNet Generalization Results}
We provide an expanded version of Table~\ref{tab:ood} that includes the anchored training protocol without the reference-masking regularizer. 
\begin{table*}[t]
\centering
\renewcommand{\arraystretch}{1.4}
\caption{\textbf{Generalization performance of models trained on ImageNet-1K}. We compare the generalization performance of different training strategies under both ID and OOD (corruptions and distribution shifts) test settings. For evaluating the prediction performance on each of the benchmarks, we consider the widely adopted Top1 accuracy metric. For calibration performance, we report the mean and standard deviation of the Smoothed ECE ($\downarrow$) metric across all ImageNet OOD datasets. Note, we highlight the best performing model in each case with \color[HTML]{B71BAA}{pink}.}
\label{tab:ood_appendix}
\resizebox{1.0\textwidth}{!}{
\begin{tabular}{c|c|c|c|c|ccccc|ccccc|c}
\toprule
 &  &  &  &  &  \multicolumn{5}{c|}{\textbf{ImageNet-C}} & \multicolumn{5}{c|}{\textbf{ImageNet-$\bar{\text{C}}$}} & \\ 
\cline{6-10}\cline{11-15}
\textbf{Model}  & \textbf{Method} & \textbf{ID Acc.} & \textbf{ImageNet-R} & \textbf{ImageNet-S} & \textbf{Sev. 1} & \textbf{Sev. 2} & \textbf{Sev. 3} & \textbf{Sev. 4} & \textbf{Sev. 5} & \textbf{Sev. 1} & \textbf{Sev. 2} & \textbf{Sev. 3} & \textbf{Sev. 4} & \textbf{Sev. 5} & \textbf{Calibration}\\ 

\midrule
 & Standard & $82.07$ & $40.84$ & $27.08$ & $71.63$ & $64.89$ & $57.77$ & $47.77$ & $35.66$ & $71.37$ & $67.12$ & $61.2$ & $52.01$ & $46.54$ & $0.121\pm0.034$\\ 

 & Anchoring & \color[HTML]{B71BAA}$\mathbf{82.26}$ & $40.36$ & $27.56$ & \color[HTML]{B71BAA}$\mathbf{72.32}$ & \color[HTML]{B71BAA}$\mathbf{65.85}$ & $58.95$ & $49.51$ & $37.41$ & $72.68$ & $68.96$ & $63.29$ & $53.74$ & $48.14$ & $0.121\pm0.032$\\ 

\multirow{-3}{*}{SWINv2-T ($28.4$M)} & Proposed & $82.03$ & \color[HTML]{B71BAA}$\mathbf{41.17}$ & \color[HTML]{B71BAA}$\mathbf{27.68}$ & $72.13$ & $65.71$ & \color[HTML]{B71BAA}$\mathbf{59.21}$ & \color[HTML]{B71BAA}$\mathbf{50.01}$ & \color[HTML]{B71BAA}$\mathbf{38.58}$ & \color[HTML]{B71BAA}$\mathbf{73.51}$ & \color[HTML]{B71BAA}$\mathbf{70.45}$ & \color[HTML]{B71BAA}$\mathbf{65.77}$ & \color[HTML]{B71BAA}$\mathbf{57.31}$ & \color[HTML]{B71BAA}$\mathbf{51.76}$ & \color[HTML]{B71BAA}$\mathbf{0.117\pm0.027}$\\ 

\midrule
 & Standard & $83.71$ & $45.17$ & $32.25$ & $74.48$ & $68.8$ & $62.84$ & $54.32$ & $42.85$ & $75.39$ & $72.26$ & $67.14$ & $58.73$ & $53.7$ & $0.126\pm0.039$\\ 

 & Anchoring & \color[HTML]{B71BAA}$\mathbf{84.0}$ &  $45.95$ & $32.08$ & \color[HTML]{B71BAA}$\mathbf{74.75}$ & $68.87$ & $63.12$ & $54.7$ & $43.14$ & $76.07$ & $73.33$ & $68.79$ & $60.49$ & $55.19$ & $0.122\pm0.045$\\ 

\multirow{-3}{*}{SWINv2-S ($49.7$M)} & Proposed & $83.68$ & \color[HTML]{B71BAA}$\mathbf{46.63}$ & \color[HTML]{B71BAA}$\mathbf{33.3}$ & $74.7$ & \color[HTML]{B71BAA}$\mathbf{69.12}$ & \color[HTML]{B71BAA}$\mathbf{63.65}$ & \color[HTML]{B71BAA}$\mathbf{55.5}$ & \color[HTML]{B71BAA}$\mathbf{44.33}$ & \color[HTML]{B71BAA}$\mathbf{76.59}$ & \color[HTML]{B71BAA}$\mathbf{74.24}$ & \color[HTML]{B71BAA}$\mathbf{70.17}$ & \color[HTML]{B71BAA}$\mathbf{62.93}$ & \color[HTML]{B71BAA}$\mathbf{58.25}$ & \color[HTML]{B71BAA}$\mathbf{0.119\pm0.041}$\\ 

\midrule
 & Standard & \color[HTML]{B71BAA}$\mathbf{81.07}$ & $44.06$ & $29.4$ & $72.37$ & $66.57$ & $61.6$ & $52.88$ & $41.09$ & $72.75$ & $69.01$ & $63.47$ & $54.7$ & $50.07$ & $0.109\pm0.037$\\ 

 & Anchoring & $80.57$ & $45.56$ & $32.32$ & \color[HTML]{B71BAA}$\mathbf{72.64}$ & $67.14$ & $62.33$ & $54.46$ & $43.48$ & $73.21$ & $69.74$ & $64.57$ & $56.03$ & $51.46$ & $0.106\pm0.035$\\ 

\multirow{-3}{*}{VITb16 ($86.6$M)} & Proposed & $80.76$ &  \color[HTML]{B71BAA}$\mathbf{46.39}$ & \color[HTML]{B71BAA}$\mathbf{33.0}$ & $72.52$ & \color[HTML]{B71BAA}$\mathbf{67.38}$ & \color[HTML]{B71BAA}$\mathbf{62.87}$ & \color[HTML]{B71BAA}$\mathbf{55.13}$ & \color[HTML]{B71BAA}$\mathbf{44.52}$ & \color[HTML]{B71BAA}$\mathbf{73.65}$ & \color[HTML]{B71BAA}$\mathbf{70.91}$ & \color[HTML]{B71BAA}$\mathbf{66.87}$ & \color[HTML]{B71BAA}$\mathbf{59.29}$ & \color[HTML]{B71BAA}$\mathbf{54.94}$ & \color[HTML]{B71BAA}$\mathbf{0.105\pm0.028}$\\ 

\midrule
 & Standard & \color[HTML]{B71BAA}$\mathbf{84.11}$ & $45.7$ & $31.91$ & $74.45$ & $68.55$ & $62.34$ & $53.66$ & $41.87$ & $75.12$ & $72.15$ & $67.16$ & $58.66$ & $53.75$ & $0.132\pm0.055$\\ 

 & Anchoring & $84.06$ & $47.6$ & \color[HTML]{B71BAA}$\mathbf{33.42}$ & $74.95$ & $69.28$ & $63.43$ & $55.08$ & $43.8$ & $76.36$ & $73.3$ & $68.49$ & $60.05$ & $54.81$ & $0.129\pm0.058$\\ 

\multirow{-3}{*}{SWINv2-B ($87.8$M)} & Proposed & $84.09$ & \color[HTML]{B71BAA}$\mathbf{48.16}$ & $33.34$ & \color[HTML]{B71BAA}$\mathbf{75.24}$ & \color[HTML]{B71BAA}$\mathbf{69.63}$ & \color[HTML]{B71BAA}$\mathbf{64.05}$ & \color[HTML]{B71BAA}$\mathbf{56.08}$ & \color[HTML]{B71BAA}$\mathbf{45.19}$ & \color[HTML]{B71BAA}$\mathbf{77.1}$ & \color[HTML]{B71BAA}$\mathbf{74.69}$ & \color[HTML]{B71BAA}$\mathbf{70.81}$ & \color[HTML]{B71BAA}$\mathbf{63.53}$ & \color[HTML]{B71BAA}$\mathbf{58.77}$ & \color[HTML]{B71BAA}$\mathbf{0.124\pm0.051}$\\ 

\midrule
\end{tabular}
}
\end{table*}

\begin{table}[t]
\centering
\renewcommand{\arraystretch}{1.4}
\caption{Measuring anomaly rejection performance on Imagenet-1K. We report the AUROC ($\uparrow$) scores}
\label{tab:anomaly_expanded}
\resizebox{0.6\textwidth}{!}{
\begin{tabular}{c|c|ccccc}
\toprule
 &     & \multicolumn{5}{c}{\textbf{Anomaly Rejection (AUROC)}} \\ 
\cline{3-7}
\textbf{Architecture} & \textbf{Method}  & \textbf{iSUN} & \textbf{Textures} & \textbf{Places365}\\ 

\midrule
 & Standard & \color[HTML]{B71BAA}$\mathbf{80.25}$ & $76.83$ & $72.53$\\ 

 & Anchored Training & $78.68$ & $76.64$ & $74.75$ \\ 

\multirow{-3}{*}{SWINv2-T} & Proposed   & $77.69$ & \color[HTML]{B71BAA}$\mathbf{78.09}$ & \color[HTML]{B71BAA}$\mathbf{77.16}$ \\ 

\midrule
 & Standard &   $82.89$ & $77.87$ & $70.63$ \\ 

 & Anchored Training   &   \color[HTML]{B71BAA}$\mathbf{87.73}$ & \color[HTML]{B71BAA}$\mathbf{80.83}$ & \color[HTML]{B71BAA}$\mathbf{76.67}$ \\ 

\multirow{-3}{*}{SWINv2-S} & Proposed   & $84.18$ & $79.66$ & $74.85$ \\ 

\midrule
 & Standard & \color[HTML]{B71BAA}$\mathbf{86.92}$ & \color[HTML]{B71BAA}$\mathbf{79.24}$ & $65.72$ \\ 

 & Anchored Training  & $85.17$ & $76.88$ & $66.16$ \\ 

\multirow{-3}{*}{VITb16} & Proposed & $84.55$ & $78.91$ & \color[HTML]{B71BAA}$\mathbf{67.18}$ \\ 

\midrule
 & Standard & $85.32$ & $76.35$ & $65.99$ \\ 

 & Anchored Training  & $85.98$ & \color[HTML]{B71BAA}$\mathbf{77.88}$ & $70.75$ \\ 

\multirow{-3}{*}{SWINv2-B} & Proposed & \color[HTML]{B71BAA}$\mathbf{87.34}$ & $75.74$ & \color[HTML]{B71BAA}$\mathbf{73.66}$ \\ 
\bottomrule
\end{tabular}
}
\end{table}

\section{Expanded Anomaly Rejection Results for Vision OOD Datasets}
While Table~\ref{tab:anomaly} in the main paper provided anomaly rejection results averaged over all Vision OOD datasets,  we expand and present metrics for each dataset in Table~\ref{tab:anomaly_expanded}

\end{document}